\documentclass[10pt,twocolumn,letterpaper]{article}

\usepackage[pagenumbers]{cvpr} 

\usepackage[dvipsnames]{xcolor}

\usepackage{graphicx}
\usepackage{amsmath}
\usepackage{amssymb}
\usepackage{booktabs}

\usepackage{multirow,booktabs, hhline}
\usepackage{amssymb}
\usepackage{pifont}
\usepackage{makecell}
\usepackage{arydshln}
\usepackage{algorithm}
\usepackage{algorithmicx}
\usepackage{algpseudocode}
\usepackage{multirow}
\usepackage{array}
\usepackage{soul}
\usepackage{xcolor}
\usepackage{color}
\usepackage{xspace}
\usepackage{tabularx}
\usepackage{wrapfig}
\usepackage{pifont}
\usepackage{stfloats}
\usepackage[misc]{ifsym}
\newcolumntype{C}[1]{>{\centering\arraybackslash}p{#1}}
\newcolumntype{L}[1]{>{\arraybackslash}p{#1}}

\newcommand{\figref}[2][{}]{\hyperref[#2]{\figurename~\ref{#2}#1}} 
\definecolor{cvprblue}{rgb}{0.21,0.49,0.74}
\usepackage[pagebackref,breaklinks,colorlinks,citecolor=cvprblue]{hyperref}

\usepackage[capitalize]{cleveref}
\crefname{section}{Sec.}{Secs.}
\Crefname{section}{Section}{Sections}
\Crefname{table}{Table}{Tables}
\crefname{table}{Tab.}{Tabs.}

\definecolor{lightergray}{RGB}{230,230,230}
\definecolor{DarkGreen}{RGB}{30,130,30}
\definecolor{red}{RGB}{255,0,0}
\newcommand{\cmark}{\textcolor{DarkGreen}{\ding{51}}}
\newcommand{\xmark}{\textcolor{red}{\ding{55}}}%

\newcommand\blfootnote[1]{%
\begingroup
\renewcommand\thefootnote{}\footnote{#1}%
\addtocounter{footnote}{-1}%
\endgroup
}

\def\paperID{2073} 
\def\confName{CVPR}
\def\confYear{2024}

\title{ControlLLM: Augment Language Models with Tools by Searching on Graphs}

\author{Zhaoyang Liu$^{*1,2}$  \quad Zeqiang Lai$^{*2}$ \quad Zhangwei Gao$^{2}$  \quad  Erfei Cui$^{2}$ \quad Ziheng Li$^{3}$ \\ Xizhou Zhu$^{2,3}$ \quad Lewei Lu{$^{4}$} \quad Qifeng Chen\textsuperscript{${1}~$\Letter} \quad Yu Qiao$^{2}$ \quad Jifeng Dai$^{2,3}$ \quad Wenhai Wang\textsuperscript{${2}~$\Letter}\\
$^{1}$The Hong Kong University of Science and Technology \\ $^{2}$OpenGVLab, Shanghai AI Laboratory \quad $^{3}$Tsinghua University \quad $^{4}$SenseTime\\
\\
{\small \url{https://github.com/OpenGVLab/ControlLLM}}
}

\begin{document}

\maketitle

\begin{abstract}
We present ControlLLM, a novel framework that enables large language models (LLMs) to utilize multi-modal tools for solving complex real-world tasks. Despite the remarkable performance of LLMs, they still struggle with tool invocation due to ambiguous user prompts, inaccurate tool selection and parameterization, and inefficient tool scheduling. To overcome these challenges, our framework comprises three key components: (1) a \textit{task decomposer} that breaks down a complex task into clear subtasks with well-defined inputs and outputs; (2) a \textit{Thoughts-on-Graph (ToG) paradigm} that searches the optimal solution path on a pre-built tool graph, which specifies the parameter and dependency relations among different tools; and (3) an \textit{execution engine with a rich toolbox} that interprets the solution path and runs the tools efficiently on different computational devices. We evaluate our framework on diverse tasks involving image, audio, and video processing, demonstrating its superior accuracy, efficiency, and versatility compared to existing methods.
\end{abstract}
\blfootnote{ *\ Equal contribution.}
\blfootnote{ \Letter\ Corresponding authors (wangwenhai@pjlab.org.cn, cqf@ust.hk).}

\section{Introduction}

\begin{figure*}[t]
\centering
\includegraphics[width=1\linewidth]{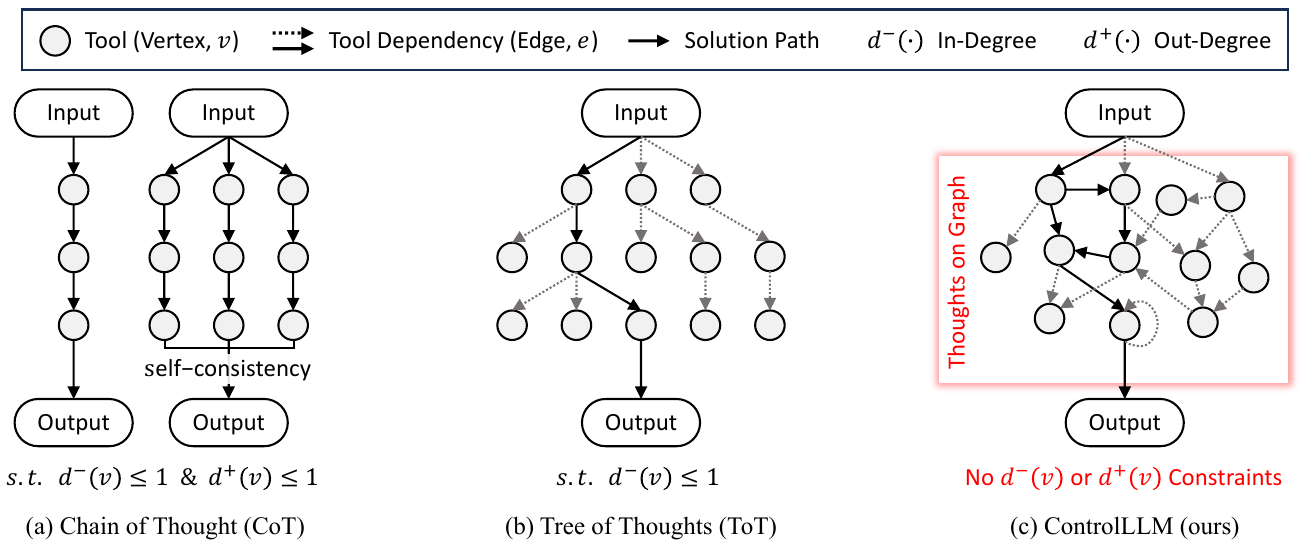}
\caption{\textbf{Comparisons of different paradigms for task planning.}
(a) Chain of Thought (CoT)~\cite{wei2022chain}, CoT with self-consistency~\cite{wang2022self} and (b) Tree of Thoughts~\cite{yao2023tree} (ToT) essentially rely on the LLMs to perform task planning, where the edge is actually formed by LLMs at run time. (c) The Thoughts-on-Graph (ToG) paradigm in our method searches for solutions on a pre-built graph that captures the dependencies of tools, which avoids the hallucination problem in tool invocation.
}
\label{fig:motivation}
\end{figure*}
Large-scale language models, such as ChatGPT~\cite{chatgpt} and LLaMA series~\cite{touvron2023llama,touvron2023llama2}, have demonstrated impressive capability in understanding and generating natural language. 
Beyond their prowess in linguistic tasks, these models have been rapidly extended to interaction, planning, and reasoning, propelling the advancement of studies in multi-modal interaction~\cite{li2022blip,li2023blip, wang2023visionllm,zhang2023llama,ma2023cephgpt,ahn2022can,vemprala2023chatgpt}.

One of the emerging examples of multi-modal interaction is tool-augmented language models~\cite{shen2023hugginggpt,wu2023visual,yang2023gpt4tools,schick2023toolformer,liu2023internchat}, which strive to enhance the capabilities of language models to include diverse modalities beyond text such as image, video, audio, \etc.
These models employ LLMs as primary controllers and incorporate tools with diverse functionalities as plugins, 
which solves a wide range of multi-modal tasks. 
However, challenges in this field still persist, covering task decomposition, task planning, and efficient tool scheduling.

With these challenges in mind, prior methods~\cite{yao2022react,shen2023hugginggpt,liu2023internchat,wu2023visual,yang2023gpt4tools,qin2023toolllm,tang2023toolalpaca} made their endeavors in developing tool-augmented LLMs. They utilize LLMs with input-output prompting, CoT~\cite{wei2022chain} or ToT~\cite{yao2023tree} to perform task planning. These methods can solve problems by breaking them into a chain or tree of sub-tasks. Theoretically, as long as LLMs have strong generalization ability, these methods can also solve complex tasks. However, in practice, we found that these methods often suffer from inaccurate tool invocation problems when dealing with complex cases. This is due to the fact that solutions for complex tasks often contain tool invocations with intricate topological structures. It is insufficient for these methods to form a complex thought network and thus fail to solve complicated tasks. 
Therefore, it requires us to figure out a new paradigm beyond chain-shaped or tree-shaped ones, which can generate solutions with intricate topology structures to solve more complicated problems (see Fig.~\ref{fig:motivation} and Fig.~\ref{fig:system_design}).

To this end, we introduce ControlLLM, a new framework that assists large language models in accurately and efficiently controlling multi-modal tools and identifying comprehensive solutions for complex real-world tasks involving multi-modal inputs. 
Alongside a variety of improvements over previous works, 
our framework places particular emphasis on three aspects as follows:

\textbf{Task Decomposition.} A task decomposer is introduced to analyze the user prompt and breaks it down into a number of subtasks, each with well-defined attributes such as task description, task domain, arguments, and returned output. By decomposing complex tasks into manageable subtasks, the task decomposer significantly enhances the system's ability to handle intricate user prompts, which paves the way for follow-up task planning and solution execution.

\textbf{Task Planning.} This part handles tool selection and tool argument assignment. We propose a thoughts-on-graph (ToG) paradigm that traverses a topological tool graph to search for solutions. The nodes of the graph are tools that are connected based on their dependencies and relationships. ToG orchestrates the selected tools and controls the flow of resources among them to form possible solutions. ToG can find the optimal solution for each sub-task by applying diverse search strategies on the graph. 
Due to the concrete definition in subtask and explicit tool dependencies in a tool graph, ToG can effectively search all feasible solution paths in cases where the selected optimal solution fails to meet users' preferences.

\textbf{Solution Execution.} We design an execution engine that can execute the solution generated by ToG and craft informative and well-formatted responses. The engine has access to a versatile toolbox consisting of various tools from different sources, such as locally deployed APIs or cloud services. The engine can also parallelize the tool executions according to the topology of the solution path to reduce the latency and provide feedback during the execution process.

Our ControlLLM offers several advantages. 
(1) It can accurately handle complex real-world tasks that involve multi-modal inputs and outputs, while previous methods~\cite{vicuna2023,liu2023visual,shen2023hugginggpt,wu2023visual,yang2023gpt4tools,liu2023internchat} usually fail to handle due to their capabilities of task planning;
(2) It can overcome the token limitation of LLMs during task planning. Because our method searches the optimal solution path on the tool graph, instead of asking LLMs to generate a solution for the task;
(3) It can easily scale up toolbox. Since all solutions lie in the tool graph, when tools change, we only need to rebuild the graph without re-training LLMs or updating in-context prompts. 

To evaluate the effectiveness of ControlLLM in tasks of different complexities, we construct a benchmark with a series of tailored metrics. Specifically, we use irrelevant tool inclusion rate and necessary tool inclusion rate to measure tool selection. We employ the resource hallucination rate and resource type consistency rate to assess argument assignments. We also split the test set into three difficulty levels based on the number of APIs involved: easy ($<2$ APIs), medium ($2$ or $3$ APIs), and hard ($>3$ APIs). We conducted various experiments, both quantitatively and qualitatively, to compare our method with existing ones. The results show that ControlLLM achieves a higher success rate in tool invocation, especially for complicated instructions.

In summary, the main contributions are as follows:

(1) We propose ControlLLM, a framework that lets LLMs use various tools across different modalities to solve complex tasks in the real world. With a powerful toolbox, ControlLLM can be easily extended to tasks with natural language, images, audio, video, or any mix of them.

(2) We design three tailored components in ControlLLM: Task decomposition, which breaks down the user prompt into subtasks with well-defined inputs and outputs; ToG paradigm for task planning, searching the optimal solution path on a graph that depicts tool dependencies; And an execution engine with a powerful toolbox, which efficiently schedules and executes the solution path.

(3) We construct a benchmark to assess the efficacy of ControlLLM on tasks with different complexity levels. The experimental results demonstrate significant improvements in tool usage. Notably, ControlLLM achieves a success rate of 93\% in the metric of overall solution evaluation on challenging tasks, while the best baseline only reaches 59\%.

\section{Related Work}
\textbf{Planning, Reasoning, and Decision Making.}
It is a longstanding vision to empower autonomous agents with the abilities of planning, reasoning, and decision-making~\cite{weiss1978model, latombe2012robot, silver2017mastering}. Despite progressive development, it was recent advancements in large language models (LLM)~\cite{brown2020language, chowdhery2022palm, touvron2023llama, ouyang2022training,zeng2022glm} that have taken a breakthrough step in addressing these problems on the broad user requests. Nevertheless, it is shown that LLMs still suffer from difficulties in dealing with knowledge-heavy and complex tasks~\cite{rae2021scaling}. To overcome these issues, Chain of Thoughts (CoT)~\cite{wei2022chain} is introduced as a simple Tool Documentation Enables Zero-Shot Tool-Usage with Large Language Modelsprompting technique to elite the complex reasoning capabilities of LLMs. Following this line of work, CoT with self consistency~\cite{wang2022self}, Tree of Thoughts (ToT)~\cite{wei2022chain}, and other techniques~\cite{zhou2022least, chung2022scaling, ho2022large}, have been proposed to improve the reasoning abilities further. 
There are also several works~\cite{yao2023beyond, besta2023got} that introduce techniques called Graph-of-Thought (GoT). 
They all share a common insight that relies on LLMs to generate thoughts for solving complicated NLP problems. In contrast, our ToG aims to endow the language model with the ability to use tools for a multi-modal dialogue system. Furthermore, ToG builds a tool-graph in advance without requiring LLMs and uses a search algorithm to form a complicated thought network for task planning.

\textbf{Tool-Augmented LLM.}
Drawing inspiration from the evolving planning and decision-making capabilities observed in Large Language Model (LLM) systems, a new wave of research starts to enhance LLMs with external tools for accessing up-to-date information, reducing hallucination, multi-modal interactions, \etc. Prominent examples include ReAct~\cite{yao2022react}, VisProg~\cite{gupta2023visual}, Visual ChatGPT~\cite{wu2023visual}, HuggingGPT~\cite{shen2023hugginggpt}, InternGPT~\cite{liu2023internchat}, AutoGPT\footnote{\url{https://github.com/Significant-Gravitas/Auto-GPT}}, and Transformers Agent\footnote{\url{https://huggingface.co/docs/transformers/transformers_agents}}. 
A distinctive trait of this line of research is its reliance on the zero-shot or few-shot in-context learning~\cite{dong2022survey} capabilities inherent in LLMs~\cite{brown2020language}. These capabilities enable task decomposition, tool selection, and parameter completion without requiring explicit finetuning. However, due to the inherent limitations of LLMs, issues such as hallucination and challenges in effective decomposition and deduction can arise with substantial frequency.
Furthermore, there are also instruction-tuning methods~\cite{parisi2022talm,schick2023toolformer,yang2023gpt4tools,hao2023toolkengpt,qin2023toolllm,patil2023gorilla}. Whereas alleviating the above issues after being tuned on the text corpus involved tools, these methods are still limited at expanding the toolset, i.e., additional training is required to add tools. Among these methods, ToolLLM~\cite{qin2023toolllm} proposes the depth first search-based decision tree to boost the planning ability of LLMs. However, it still has limitations similar to ToT, as shown in the Fig.~\ref{fig:motivation}.

\textbf{Multi-Modal LLMs.}
Developing LLMs that inherently possess multi-modal capabilities is another approach to extending the usage boundary of LLMs for more complex real-world scenarios~\cite{liu2023visual,koh2023generating,peng2023kosmos,dai2023instructblip,wu2023next,moon2023anymal,mu2023embodiedgpt,liu2023llava}. For instance, BLIP-2~\cite{li2023blip}, LLava~\cite{liu2023llava}, and Mini-GPT4~\cite{zheng2023minigpt5} bind frozen image encoders and LLMs to enable the vision-language understanding and generation. 
Similarly, VisionLLM~\cite{wang2023visionllm} and LISA~\cite{lai2023lisa} empower the LLMs with the visual perception capabilities such as object detection and segmentation. 
GILL~\cite{koh2023generating}, DreamLLM~\cite{dong2023dreamllm}, and Mini-GPT5~\cite{zheng2023minigpt} extend LLM for interleaved image and text generation by jointly optimizing the LLM with off-the-shelf Stable Diffusion model. 
Kosmos2~\cite{peng2023kosmos}, Ferret~\cite{you2023ferret}, GPT4RoI~\cite{zhang2023gpt4roi}, and \etc, design various region-aware image encoders to augment LLMs with the abilities of grounding and referring. 
Nevertheless, these methods could only cover a limited range of modalities or tasks and often require huge effects on model finetuning.

\section{ControlLLM}
\begin{figure*}[t]
\centering
\includegraphics[width=1\linewidth]{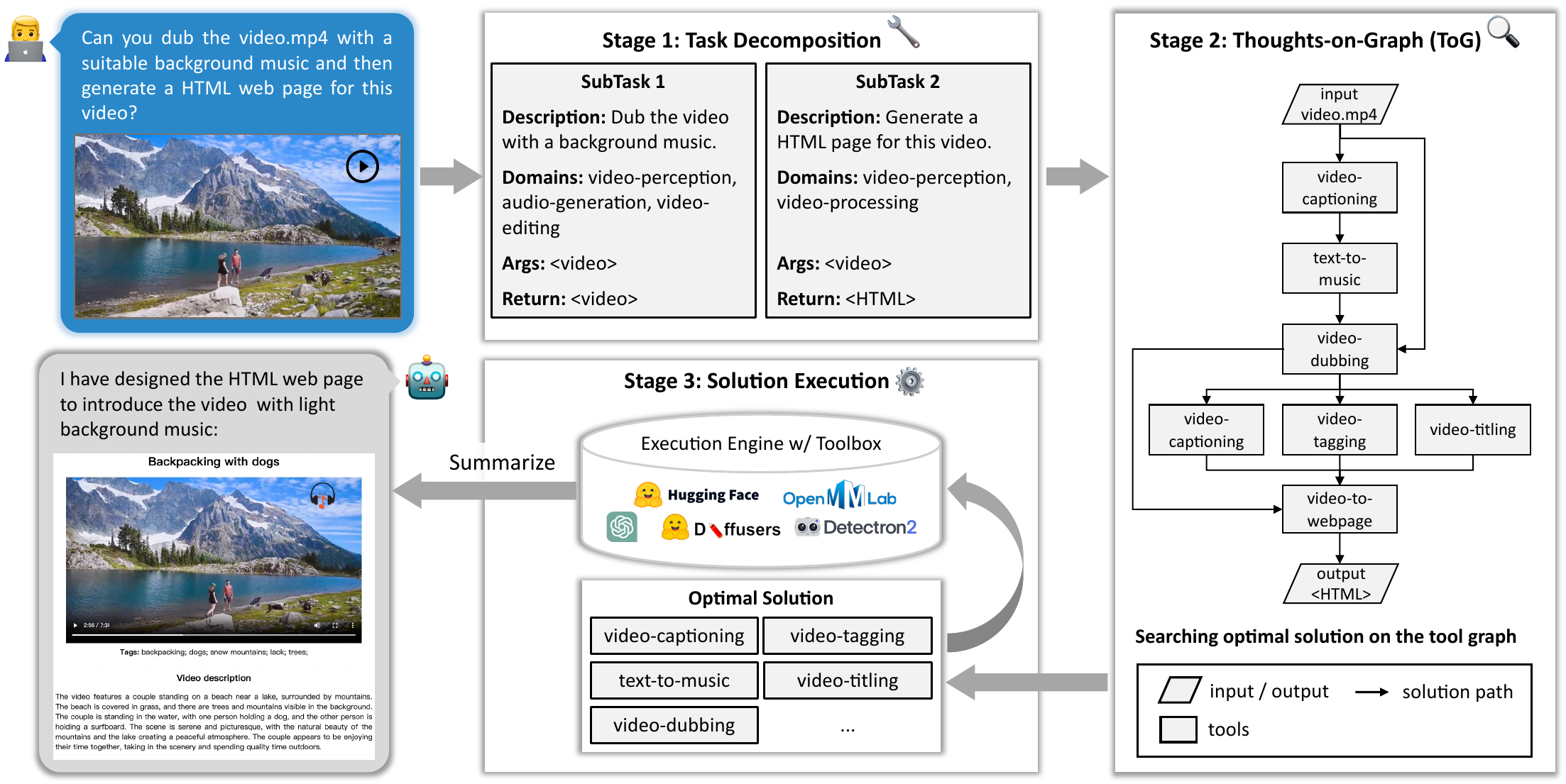}
\caption{\textbf{System design of ControlLLM.} The framework consists of three stages. The first stage is task decomposition, which parses the user input into several subtasks. Then, in Stage 2, ToG utilizes a depth-first search algorithm to find the optimal solution for each subtask. The execution engine in the last stage executes the solution and returns the output to users. We here use the example of generating a web page for the video to illustrate our method.}
\label{fig:system_design}
\end{figure*}

The prevalence of LLMs has unprecedentedly boosted the development of human-computer interaction. It is feasible to empower the LLMs with abilities to interact with various modalities via tools. In response, we present an innovative framework, namely \textbf{ControlLLM}, characterized by its flexibility, and high performance. 
As depicted in \cref{fig:system_design}, our framework consists of three sequential stages, \ie, task decomposition, task planning and solution execution. Next, we will illustrate the design of each stage in detail.

\subsection{Task Decomposition}

ControlLLM starts with task decomposition -- a stage for decomposing the user request $r$ into a list of parallel subtasks. We here can utilize a language model $\mathcal{M}$, \eg, ChatGPT or instruction-tuned LLaMA, to automatically decompose the user request as follows: 
\begin{equation}
\label{eqn:task_decomposition}
    \begin{aligned}
        \{s_0, ..., s_i ,..., s_n\} = \mathcal{M}(r),
    \end{aligned}
\end{equation}
where $s_i$ is the i-th subtask, $n$ is the number of all subtasks. We will elaborate on the different choices of language model $\mathcal{M}$ in Sec.~\ref{sec:lm} and discuss their impacts in Sec.~\ref{sec:ablation}.
The result of task decomposition is JSON format, and the output protocol is presented in Table~\ref{tab:decomposition_protocol}.

Task decomposition is different from task planning. It only breaks down the user's request into several parallel subtasks and summarizes the input resources for each subtask from the user request. It does not need to know what tools to use or how to use them.
The objective of this stage is to achieve three aims. Firstly, it splits user requests into smaller and more manageable units, \ie, subtasks, thereby accelerating task planning. Secondly, it seeks to determine the task domain that is most relevant and appropriate for the given problem, thus further narrowing down the scope of task planning. Thirdly, it endeavors to infer the input and output resource types from the context, which identifies the start and end nodes for ToG to search in the next stage.

\subsection{Task Planning with Thoughts-on-Graph} 
This stage is the key of the entire system. Given the results of task decomposition, we design a Thoughts-on-Graph (ToG) paradigm to find solutions on the graph heuristically. 

\subsubsection{Building the Tool Graph}
\label{sec_build_graph}
In this stage, we embark on constructing a Tool Graph $G$ by simply using an adjacency matrix, which serves as a fundamental guideline for analyzing and optimizing the interactions between tools. Our motivation is driven by observing a discernible topological structure that inherently exists between the input and output of diverse tools, as demonstrated in Fig.~\ref{fig:system_design}. This compelling insight propels us to craft a comprehensive tool graph that encapsulates the inherent relationship between tools.

There are two types of nodes \ie, \textit{Resource} node and \textit{Tool} node, in the graph. \textit{Resource} node can be formally defined as one-tuple: $\left<\text{\texttt{type}}\right>$, where \texttt{type} represents the specific type of resource, like image, mask, video, \etc. 
\textit{Tool} node can be expressed as a three-tuple: $
\left<\text{\texttt{desc, args, ret}}\right>$. The \texttt{desc} field encapsulates the description of the tool, elucidating its purpose, methodology, and intended applications. The \texttt{args} field represents a list of resource nodes that the tool accepts, thereby giving the prerequisites for utilizing this tool. Finally, the \texttt{ret} field designates the resource node that the tool returns. We elaborate on the definitions of resource types and tools in supplementary material (Sec.~\ref{sec:tool_type}).

\textbf{Edge Definitions.}
Edges in the tool graph intricately connect the nodes, highlighting the relationships between different tools. We define two types of edges in the graph.

(1) \textit{Tool-resource edge} is established from the tool to its returned resource type. This signifies that the tool is capable of generating resources of the corresponding type. Mathematically, a tool-resource edge is represented as:

\begin{equation}
G(T_j, R_i)=\left\{
\begin{aligned}
\text{true } & , & \text{if R$_i$ equals to   \texttt{ret} of T$_j$} \\
\text{false} & , & \text{otherwise}
\end{aligned}
\right.,
\end{equation}
where $T_j$ is $j$-th tool node, $R_i$ is $i$-th resource node, ``true" denotes two nodes are connected, and ``false" denotes two nodes are disconnected.

(2) \textit{Resource-tool edge} denotes the resource node that can be accepted as input arguments for its adjacent tool. This connection indicates how the resources flow to the tool. The resource-tool edge is mathematically defined as:
\begin{equation}
G(R_i, T_j)=\left\{
\begin{aligned}
\text{true } & , & \text{if $R_i$ belongs to \texttt{args} of  $T_j$} \\
\text{false} & , & \text{otherwise}
\end{aligned}
\right..
\end{equation}

Through the establishment of this graph, we can use diverse search strategies to make informed decisions regarding tool selection, 
 and input resource assignments.

\subsubsection{Searching on the Graph}
\label{sec:search_on_the_graph}

\definecolor{dg}{rgb}{0.0, 0.5, 0.0}
\begin{algorithm}[t]
\caption{The Python pseudocode of depth-first solution search in Thoughts-on-Graph}
\label{algo:got}
\begin{algorithmic}[1]
{{

\Require ~~{\newline t: subtask obtained by Eq.~\ref{eqn:task_decomposition}\newline
g: tool graph $G$ constructed in Sec.~\ref{sec_build_graph} \newline
r: available resources, initialized with subtask[“args”]\newline
s: recorded tools during searching}
\Ensure ~~{\newline solutions: all possible solutions for the subtask t}
\Function{DFS\_Search}{t, g, r, s}
    \State\textbf{if} {len(s) $>$ $m$}:
        \State \quad \Return{[]}
    \newline\textcolor{dg}{\quad\ \ \# $\mathcal{F}$ finds all tool candidates, explained in Sec.~\ref{sec:search_on_the_graph}}
    \State available\_tools = $\mathcal{F}$(t, g, r)
    \State solutions = []
    \State\textbf{for} tool \textbf{in} available\_tools:
        \State \quad s.append(tool)
        \State \quad r.append(tool[``returns''])
        \State\quad\textbf{if} {tool[``returns"] == t[``returns"]}:
            \State\qquad solutions.append(s.copy())
        \State\quad results = \Call{DFS\_Search} {t, g, r, s}
        \State\quad solutions.extend(results)
        \State\quad r.remove(tool[``returns"])
        \State\quad s.remove(tool)
    \State \Return{solutions} $ \hfill \triangleright {\rm \ Return}$
\EndFunction

}
}
\end{algorithmic}
\end{algorithm}

As described in Algorithm~\ref{algo:got}, our ToG is built upon a depth-first search (\textit{DFS}) algorithm where the tool selection function $\mathcal{F}$ is used to sample the tool nodes on the tool graph. The algorithm starts from the input resource nodes and explores all possible paths to the output node while keeping track of the intermediate resources and tools along the way. The algorithm stops when it reaches the expected output node or when it exceeds a maximum length limit $m$ ($m$=$10$ by default). Finally, the algorithm returns all searched solutions as a list of tool sequences. Each step from \textit{resource node} to \textit{tool node} represents a thought process, as it involves a decision that determines whether to use this tool and how to assign its input arguments from available resources.

To find a trade-off between time and space complexities, we develop a tool assessment module in which the language model is leveraged to score the tools in each search step and then filter out some irrelevant tools. For details, please refer to Sec.~\ref{sec:tool_assessment} in supplementary material. With this assessment module, we design four search strategies for the function $\mathcal{F}$ to determine which tool nodes within the task domains to visit among all adjacent nodes when searching on the graph:

\textbf{Greedy Strategy.} This strategy selects the tool node with the highest score at each step, where the score indicates the relevance of the tool to the task. A higher score indicates that the tool is more helpful for solving the task. Greedy search is fast and simple, but it may not find the optimal solution or even any solution at all.

\textbf{Beam Strategy.} It only keeps the $k$ best tools according to their assessment scores. Beam search can expand the search space but reduce the search efficiency slightly.

\textbf{Adaptive Strategy.} This is a variant of beam search where it dynamically adjusts the beam size by choosing the tools with scores higher than a fixed threshold, which is a trade-off between exploration and exploitation. It can widen the search space when there are many available choices and narrow down when there are few confident choices.

\textbf{Exhaustive Strategy.} This strategy explores all possible paths from the start node to the terminal node. The exhaustive search is guaranteed to find an optimal solution if one exists, but it may be very slow and consume a lot of computational resources during the search.

The impacts of different search strategies are studied in  Sec.~\ref{sec:ablation}. By initiating a systematic traversal of tool graph, commencing at the “args” nodes and culminating at the “return” node, a diverse list of conceivable solutions is meticulously synthesized. This process, akin to a brainstorm or mind map, represents the spectrum of potential solutions. 

\subsubsection{Solutions Post-processing}
After ToG searches the solutions, we design \textit{solution expert} and \textit{resource expert} to post-process solutions, which both build upon the language model $\mathcal{M}$. Specifically, \textit{solution expert} to select the optimal one among all solution candidates and \textit{resource expert} to infer the remaining arguments for tools, respectively. The overall details are shown in supplementary material (Sec.~\ref{sec:solution_expert} and \ref{sec:resource_expert}).

\subsection{Solution Execution}
\label{solution_execution}

Once the task solutions are completed, they are passed to the \textit{execution engine}  obtain results, as shown in \cref{fig:system_design}.
During this stage, the execution engine initially parses the solutions into a sequence of \emph{Actions}. Each action is associated with particular tool services, which could be implemented via either handcrafted mapping tables or an automatic scheduler based on some strategies.
Different from previous works~\cite{wu2023visual,yang2023gpt4tools, liu2023internchat} that adopt static tool mapping, our design empowers the system with the flexibility to schedule diverse tools based on users' preferences.

The parsed actions are automatically executed by scheduling the action to the local, remote, or hybrid endpoints. Multiple independent subtasks would be executed in parallel to improve efficiency. Besides, we maintain a state memory storing all the intermediate results, including their values and types. This enables the running-time automatic correction for the action parameters. 

\textbf{Response Generation.}
With all the execution results in hand, we could respond to the user requests. The unprocessed results may lack comprehensiveness and clarity, potentially making it difficult for users to understand. To this end, we introduce a module to aggregate all the execution results and generate user-friendly responses. This is achieved by prompting the LLMs, such as ChatGPT, with the user request, action list, and execution results and asking them to summarize the answers intelligently. The prompt can be found in supplementary material (Table~\ref{tab:prompt_response}).

\subsection{The Choices of Language Model} 
\label{sec:lm}

One feasible yet direct choice is to use off-the-shelf \textbf{large language models} (LLMs) such as ChatGPT or Llama 2~\cite{touvron2023llama2}, which are pre-trained on large-scale text corpora and can handle various NLP tasks. These LLMs are readily available. We design a series of elaborate prompts as shown in Sec.~\ref{sec:cllm_chatgpt} for task decomposition, tool assessment, solution expert, and resource expert. We call this variant as ControlLLM-ChatGPT. In this way, we avoid training a language model from scratch. However, they may lead to low performance as they are not trained for our requirements.

The alternative choice of $\mathcal{M}$, termed as ControlLLM-LLaMA, is to finetune a language model (\eg, LLaMA~\cite{touvron2023llama}) by using self-instruct method~\cite{wang2022self}. More details of optimizing $\mathcal{M}$ can be referred to Sec.~\ref{sec:cllm_llama} in supplementary material. The advantage of this variant is that it can achieve high performance by adapting to the data and the task. Nevertheless, it requires lots of GPUs to train the model and may suffer from overfitting issues.

Regarding these choices, it is essential to carefully consider the trade-offs between readily available off-the-shelf LLMs with zero-shot capabilities and the potential for finetuning a model to achieve superior performance at the cost of computational resources. We will thus further discuss the impacts of different language models $\mathcal{M}$ in  Sec.~\ref{quantitative_comparisons} and explore the optimal settings for our framework.

\section{Experiments}
\subsection{Benchmark}
We build a benchmark that is used to evaluate our proposed framework compared with other state-of-the-art methods. In order to make fair comparisons, we only evaluate and test on the intersection of toolsets from different methods~\cite{shen2023hugginggpt,wu2023visual,liu2023internchat,yang2023gpt4tools}, all of which share comparable toolsets. The benchmark consists of a set of tasks that require various tools to solve complex problems collaboratively. It is designed to cover different task domains, such as question answering, image generation, image editing, image perception, visual question answering, \etc. In this benchmark, the tasks involve more than 20 tools across different modalities.

This benchmark includes about 100 instructions which are classified into three levels of difficulty: easy ($<2$ APIs), medium ($2$ or $3$ APIs), and hard($>3$APIs). We use test instructions with various levels to meticulously validate the ability of different methods. 
We believe that this benchmark can provide a comprehensive comparison of the tool control capabilities of different methods. In Table~\ref{tab:instructions}, we showcase some instruction samples from our benchmark. It is noticeable that there is no absolute relationship between difficulty and length of instruction.

\subsection{Evaluation Protocol}
Effectively evaluating the performance of tool-augmented LLMs remains a challenging task. The challenges stems from several factors, including the inherent ambiguities in natural language, the absence of shared benchmarks, and formatted solutions for systematically assessing different methods. Consequently, existing methods~\cite{wu2023visual,yang2023gpt4tools,shen2023hugginggpt,liu2023internchat} provide extensive case studies to validate the performance.

We found the APIs of tools in different methods are slightly inconsistent. It is hard to annotate all feasible solutions for each method. As such, we adopt an evaluation protocol via a multi-person voting approach with three annotation experts. The protocol breaks down the evaluation into three main aspects: tool selection, argument assignment, and overall solution evaluation. Please note that the evaluation protocol is independent of the tools' capabilities. When the tools and their input arguments are correct, we do not account for the case where the output fails to satisfy the user's expectations due to the limitations of tools.

\textbf{Metrics for Tool Selection:} A) Irrelevant Tool Inclusion Rate (\textit{abbr.} $IR$): This metric gauges the performance of the method in excluding irrelevant tools. It measures the proportion of the predicted solutions that contain the irrelevant tools. A higher $IR$ indicates that the method tends to include more unnecessary tools, potentially hindering effective task planning; B) Necessary Tool Inclusion Rate (\textit{abbr.} $NR$): This metric assesses the inclusion of necessary tools in the predicted solution but without considering whether the arguments of tools are correct. If $NR$ is high, it indicates the method has strong capabilities in tool selection.

\textbf{Metrics for Argument Assignment:} A) Resource Hallucination Rate (\textit{abbr.} $HR$): This indicator reveals the extent of hallucination in the method's responses when inferring the arguments for tools. It measures whether all arguments of the tools used in the predicted solution exist physically. A lower $HR$ suggests that the method is less prone to generating hallucinated content.
B) Resource Type Consistency Rate (\textit{abbr.} $CR$): This metric examines whether the types of input resources in the predicted solution match those of the corresponding tools. It evaluates the method's ability to ensure consistency of input types of tools.

\textbf{Solution Evaluation} (\textit{abbr.} $SE$) measures the success rate of all generated solutions on our benchmark. It only considers whether the output solution can effectively address the user's problem, irrespective of whether it contains irrelevant tools. A higher score in the solution evaluation indicates a stronger capability of task planning.

In summary, these intuitive metrics together provide a comprehensive assessment of tool-augmented LLMs. The formal definitions of these metrics can refer to Sec.~\ref{sec:metric}.

\begin{table*}[!t]
    \centering
    \begin{tabular}{l|ccccc}
        \Xhline{1.5pt}
        \textbf{Features} & \makecell{\textbf{ControlLLM}\\(our work)}& \makecell{ \textbf{HuggingGPT} \\\cite{shen2023hugginggpt} } & 
        \makecell{ \textbf{Visual ChatGPT}\\\cite{wu2023visual} } &  \makecell{ \textbf{InternGPT} \\\cite{liu2023internchat} } & \makecell{ \textbf{GPT4Tools~}\\\cite{yang2023gpt4tools} } 
        \\
         \Xhline{1pt}
          Image Perception   & \cmark & \cmark & \cmark & \cmark & \cmark  \\ 
          Image Editing   & \cmark & \cmark & \cmark & \cmark & \cmark  \\ 
          Image Generation   & \cmark & \cmark & \cmark & \cmark & \cmark \\ 
          Video Perception & \cmark & \cmark & \xmark & \cmark & \xmark \\ 
          Video Editing & \cmark & \cmark & \xmark & \cmark & \xmark \\ 
          Video Generation & \cmark & \cmark & \xmark & \cmark & \xmark  \\ 
          Audio Perception & \cmark & \cmark & \xmark & \xmark & \xmark \\
          Audio Generation & \cmark & \cmark & \xmark & \xmark & \xmark \\ 
          Multi-Solution& \cmark & \xmark & \xmark & \xmark & \xmark  \\ 
          Pointing Device & \cmark & \xmark & \xmark & \cmark & \xmark \\ 
          Resource Type Awareness & \cmark & \xmark & \xmark & \xmark & \xmark  \\
         \Xhline{1.5pt}
    \end{tabular}
    \caption{
    \small{\textbf{Comparisons of features between different methods.} The table shows that our framework supports more features that facilitate the user experience of multi-modal interaction. It proves the high scalability of our framework.}
    }
    \label{tab:feature_comparison}
\end{table*}
\subsection{Feature Comparisons}
Table \ref{tab:feature_comparison} presents a comprehensive feature comparison among various methods~\cite{shen2023hugginggpt,wu2023visual,liu2023internchat,yang2023gpt4tools}, highlighting ControlLLM's distinct advantages in the landscape of multi-modal interaction. Notably, ``Multi-Solution" signifies the method's ability to provide multiple feasible solutions, granting users more options. ``Pointing Device" signifies support for pointing devices such as the mouse, to enhance user experience. ``Resource Type Awareness" indicates the method's capability to discern the type of resource in the context, ensuring more context-aware responses.
In summary, ControlLLM emerges as the standout choice, excelling in various features. It offers a comprehensive set of tools in the domains of image, video, and audio. Moreover, its support for resource type awareness, multiple solutions, and pointing inputs demonstrates its adaptability and scalability, making it the highly versatile framework for diverse multi-modal interaction scenarios.

\subsection{Quantitative Comparisons}
\label{quantitative_comparisons}
\begin{table*}[!t]
\centering
\renewcommand{\arraystretch}{1.2}
\begin{tabular}{l|cc|cc|cccc}
\Xhline{1.5pt}
\multicolumn{1}{c|}{\multirow{2}{*}{Methods}} & \multicolumn{2}{c|}{Tool} & \multicolumn{2}{c|}{Argument} & \multicolumn{4}{c}{\multirow{1}{*}{Solution Evaluation~${\color[RGB]{50, 200, 50}\uparrow}$}}\\ 
 & $IR~\color{red}{\downarrow}$ & $NR~\color[RGB]{50, 200, 50}{\uparrow}$ & $HR~\color{red}{\downarrow}$ & $CR~\color[RGB]{50, 200, 50}{\uparrow}$ & All & Easy & Medium & Hard\\
 \Xhline{1pt}
HuggingGPT~\cite{shen2023hugginggpt} & $0.45$ & $0.64$ & $0.16$ & $0.69$ & $0.59$ & $0.73$ & $0.50$ & $0.33$\\
Visual ChatGPT~\cite{wu2023visual} & $0.26$ & $0.58$ & $0.09$ & $0.76$& $0.57$ & $0.73$ & $0.63$ & $0.10$ \\
InternGPT~\cite{liu2023internchat} & $0.12$ & $0.51$ & $0.49$ & $0.43$ & $ 0.44$ & $0.60$ & $0.46$ & $0.00$ \\ 
GPT4Tools~\cite{yang2023gpt4tools} & $0.19$ & $0.44$ & $0.28$ & $0.72$ & $0.43$ & $0.64$ & $0.33$ & $0.00$\\ \hline
ControlLLM-ChatGPT& $0.16$ & $0.63$ & $0.83$ & $0.83$ & $0.64$ & $0.71$ & $0.67$ & $0.43$\\
ControlLLM-LLaMA& $0.06$ & $\textbf{0.95}$ & $0.02$ & $0.98$ & $0.91$ & $0.98$ & $0.88$ & $0.76$ \\
ControlLLM-Mix$^\ast$ & $\textbf{0.03}$ & $0.93$ & $\textbf{0.02}$ & $\textbf{0.98}$ & $\textbf{0.93}$ & $\textbf{0.98}$ & $\textbf{0.96}$ & $\textbf{0.81}$\\ 
\Xhline{1.5pt}
\end{tabular}
\caption{
\small{\textbf{Comparisons with the state-of-the-art methods.} $\color{red}{\downarrow}$ means the smaller the better, $\color[RGB]{50, 200, 50}{\uparrow}$ means the larger the better. The results of state-of-the-art methods~\cite{shen2023hugginggpt, wu2023visual,liu2023internchat, yang2023gpt4tools} are reproduced on our own benchmark. $\ast$ denotes the default setting of ControlLLM if not stated.} 
}
\label{tab:sota}
\end{table*}
\begin{table*}[!t]
\centering
\renewcommand{\arraystretch}{1.2}
\begin{tabular}{l|cc|cc|cccc|c}
\Xhline{1.5pt}
\multicolumn{1}{c|}{\multirow{2}{*}{\makecell[c]{Search \\ Strategies}}} & \multicolumn{2}{c|}{Tool} & \multicolumn{2}{c|}{Argument} & \multicolumn{4}{c|}{\multirow{1}{*}{Solution Evaluation~${\color[RGB]{50, 200, 50}\uparrow}$}} & \multicolumn{1}{c}{\multirow{2}{*}{\makecell[c]{Time \\ Complexities}}}\\
& $IR$~$\color{red}{\downarrow}$ & $NR$~$\color[RGB]{50, 200, 50}{\uparrow}$ & $HR$~$\color{red}{\downarrow}$ & $CR$~$\color[RGB]{50, 200, 50}{\uparrow}$ & All & Easy & Meduim & Hard \\ 
\Xhline{1pt}
Greedy & $0.19$ & $0.49$ & $0.24$ & $0.76$ & $0.49$ & $0.56$ & $0.58$ & $0.19$ & $4.07$\\
Beam ($k=3$) & $0.14$ & $0.88$ & $0.01$ & $0.99$ & $0.88$ & $0.96$ & $0.79$ & $0.76$ & $121.29$\\
Adaptive & $\textbf{0.03}$ & $0.93$ & $0.02$ & $0.98$ & $0.93$ & $0.98$ & $0.96$ & $0.81$ & $236.49$ \\ 
Exhaustive & $0.06$ & $\textbf{0.97}$ & $\textbf{0.01}$ & $\textbf{0.99}$ & $\textbf{0.97}$ & $\textbf{1.00}$ & $\textbf{0.96}$ & $\textbf{0.91}$ & $3444.23$\\
\Xhline{1.5pt}
\end{tabular}
\caption{
\textbf{The evaluation for different search strategies.} As introduced in Sec.~\ref{sec:search_on_the_graph}, although exhaustive strategy achieves the best performance on most metrics, the adaptive strategy strikes a good balance between efficiency and effectiveness. We count the average number of visited tools to denote the time complexities for different search strategies.}

\label{tab:search}
\end{table*}

\begin{table*}[!tbh]
\centering
\renewcommand{\arraystretch}{1.2}
\begin{tabular}{l|l|cc|cc|cccc}
\Xhline{1.5pt}
\multicolumn{1}{c|}{\multirow{2}{*}{\makecell[c]{Task \\Decomp.}}} &\multicolumn{1}{c|}{\multirow{2}{*}{LLMs}} & \multicolumn{2}{c|}{Tool} & \multicolumn{2}{c|}{Argument} & \multicolumn{4}{c}{\multirow{1}{*}{Solution Evaluation~${\color[RGB]{50, 200, 50}\uparrow}$}}\\
& & $IR$~$\color{red}{\downarrow}$ & $NR$~$\color[RGB]{50, 200, 50}{\uparrow}$ & $HR$~$\color{red}{\downarrow}$ & $CR$~$\color[RGB]{50, 200, 50}{\uparrow}$ & All & Easy & Meduim & Hard \\ 
\Xhline{1pt}
\multirow{3}{*}{\makecell[c]{\textit{w/o PK}}}&Llama2-13B& $0.28$ & $0.71$ & $0.01$ & $0.99$ & $0.68$ & $0.87$ & $0.50$ & $0.38$\\
&ChatGPT-3.5~& $0.13$ & $0.84$ & $0.01$ & $0.99$ & $0.83$ & $0.99$ & $0.67$ & $0.57$\\
&GPT-4& $0.06$ & $0.91$ & $0.03$ & $0.97$ & $0.91$ & $0.98$ & $0.83$ & $0.81$ \\ \hline
\multirow{3}{*}{\textit{w/ PK}} & Llama2-13B& $0.12$ & $0.83$ & $0.04$ & $0.95$ & $0.82$ & $0.95$ & $0.71$ & $0.62$ \\
&ChatGPT-3.5& $0.03$ & $0.93$ & $0.02$ & $0.98$ & $0.93$ & $0.98$ & $0.96$ & $0.81$ \\
&GPT-4& $\textbf{0.01}$ & $\textbf{0.98}$ & $\textbf{0.02}$ & $\textbf{0.98}$ & $\textbf{0.98}$ & $\textbf{1.00}$ & $\textbf{1.00}$ & $\textbf{0.91}$\\ 
\Xhline{1.5pt}
\end{tabular}
\caption{
\textbf{The effects of task decomposition with regard to different LLMs.} \textit{PK} denotes ``prior knowledge".
We find, if adding prior knowledge, such as which tools might be used, into the subtask description, the performance of task planning can be evidently improved. 
}
\label{tab:ablation_task_decomposition}
\end{table*}

In this section, we give a comprehensive analysis of ControlLLM to compare with state-of-the-art methods, as summarized in Table \ref{tab:sota}. We provide three implementations in supplementary materials for our method: a) ControlLLM-ChatGPT leverages the ChatGPT-3.5 as language model $\mathcal{M}$; b) ControlLLM-LLaMA that finetunes a LLaMA-7B as a language model $\mathcal{M}$;
c) ControlLLM-Mix is regarded as our default setting, which finetunes LLaMA-7B as a task decomposer in the first stage while the remaining modules employ the ChatGPT to finish the tasks. ControlLLM-Mix combines the advantages of the other two variants and is abbreviated as ControlLLM in the following sections. 

Our evaluation is based on a set of metrics assessing effectiveness of task planning. ControlLLM excels in several key aspects. Notably, it achieves the lowest Irrelevant Tool Inclusion Rate ($IR$) as well as the highest Necessary Tool Inclusion Rate, indicating its ability in effective yet efficient task planning.
Furthermore, ControlLLM demonstrates superior performance in argument assignment, with the lowest Argument Hallucination Rate ($HR$) of $0.02$ and the highest Argument Type Consistency Rate ($CR$) of $0.98$. These results underscore its ability to generate accurate and consistent arguments, addressing a challenge in tool-augmented LLMs.
In the solution evaluation, ControlLLM maintains its lead with a score of $0.93$, indicating its effectiveness in resolving user requests. 
In summary, ControlLLM exhibits remarkable performance in all proposed metrics, evidently outperforming the state-of-the-art methods in this field.

\subsection{Ablation Studies}
\label{sec:ablation}
Table \ref{tab:search} investigates the impact of different search strategies within our Thoughts-on-Graph. We observe that the exhaustive search strategy outperforms others in most metrics, but this strategy is time-consuming. On the other hand, the greedy search strategy achieves the lowest performance. Because it can not search for a feasible path based on the tool with a high score due to inaccurate tool assessment. It thus usually fails to find the solution, especially in complicated cases. In addition, the adaptive strategy strikes a balance between performance metrics and time complexities, offering competitive results in most aspects. To trade-off between time and accuracy, we thus choose the adaptive strategy as our default setting.

In Table \ref{tab:ablation_task_decomposition}, we conduct ablation studies to evaluate the impact of different LLMs on task planning for ControlLLM-Mix. 
We find language models plays a decisive role in tool selection. The more powerful the language model, the higher the score of solution evaluation. Furthermore, we investigate the effects of incorporating prior knowledge into the subtask descriptions. The method without prior knowledge usually directly uses the user's request as a subtask description and does not offer any hints or suggestions on tool selections in the subtask description. In contrast, in the variant with prior knowledge, we add prior knowledge into the subtask description. The prior knowledge indeed improves the necessary tool inclusion rate ($NR$) and reduces the chance of selecting irrelevant tools ($IR$) when using the same large language model. 

\subsection{Qualitative Analyses}
\begin{figure*}[t]
\centering
\includegraphics[width=0.9\linewidth]{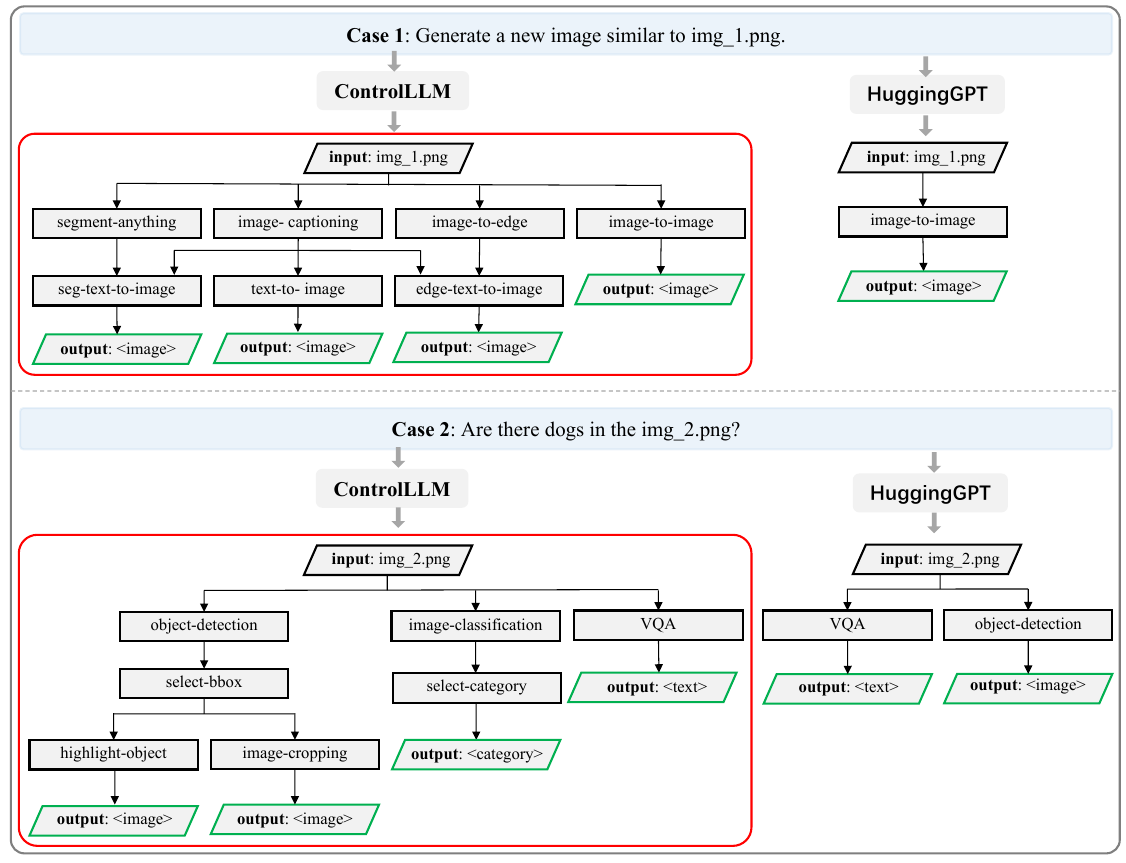}
\caption{\textbf{Qualitative comparisons of task planning.} We here use two simple cases to illustrate the differences between two different methods in task planning. Here, each output node is generated by different solution paths.}
\label{fig:solution_sample}
\end{figure*}

Fig.~\ref{fig:solution_sample} shows two simple cases to illustrate the capabilities of our ControlLLM in task planning. In contrast to HuggingGPT~\cite{shen2023hugginggpt}, we find our method is able to generate more diverse solutions to meet users' expectations, thanks to the Thoughts-on-Graph paradigm. In Sec.~\ref{sec:case_studies}, we provide extensive case studies across different modalities to validate the user experience for our method in practice.

\section{Conclusion}
In this paper, we propose \textbf{ControlLLM}, a multi-modal interaction framework that can accurately control tool usage across various domains, including text, image, audio, video, \etc. The proposed framework consists of three key stages: (1) \textit{task decomposition} to concrete the objective of the task, (2) a \textit{Thoughts-on-Graph} (ToG) paradigm to search the optimal solution path on the constructed tool graph, (3) and an \textit{execution engine} with a versatile toolbox to execute solution efficiently. 
We conduct extensive experiments and demonstrate that our ControlLLM achieves superior performance regarding tool selection, argument assignment, and overall solution effectiveness compared to existing methods. 

Nevertheless, this work still has some limitations. Since the goal of this work is to improve the accuracy of tool usage, even if the solution is theoretically feasible, we cannot guarantee that the output from tools is always correct. On the other hand, due to the inherent ambiguity in natural language, it is difficult to ensure that the optimal solution selected is consistent with the user’s goal. In this case, we can only provide more alternative solutions searched by ToG for users to choose from if the optimal solution fails.

{\small
\bibliographystyle{ieeenat_fullname}
\bibliography{egbib}

\begin{thebibliography}{55}
\providecommand{\natexlab}[1]{#1}
\providecommand{\url}[1]{\texttt{#1}}
\expandafter\ifx\csname urlstyle\endcsname\relax
  \providecommand{\doi}[1]{doi: #1}\else
  \providecommand{\doi}{doi: \begingroup \urlstyle{rm}\Url}\fi

\bibitem[Ahn et~al.(2022)Ahn, Brohan, Brown, Chebotar, Cortes, David, Finn, Fu,
  Gopalakrishnan, Hausman, et~al.]{ahn2022can}
Michael Ahn, Anthony Brohan, Noah Brown, Yevgen Chebotar, Omar Cortes, Byron
  David, Chelsea Finn, Chuyuan Fu, Keerthana Gopalakrishnan, Karol Hausman,
  et~al.
\newblock Do as i can, not as i say: Grounding language in robotic affordances.
\newblock \emph{arXiv preprint arXiv:2204.01691}, 2022.

\bibitem[Besta et~al.(2023)Besta, Blach, Kubicek, Gerstenberger, Gianinazzi,
  Gajda, Lehmann, Podstawski, Niewiadomski, Nyczyk, and Hoefler]{besta2023got}
Maciej Besta, Nils Blach, Ales Kubicek, Robert Gerstenberger, Lukas Gianinazzi,
  Joanna Gajda, Tomasz Lehmann, Micha{\l} Podstawski, Hubert Niewiadomski,
  Piotr Nyczyk, and Torsten Hoefler.
\newblock {Graph of Thoughts: Solving Elaborate Problems with Large Language
  Models}, 2023.

\bibitem[Brown et~al.(2020)Brown, Mann, Ryder, Subbiah, Kaplan, Dhariwal,
  Neelakantan, Shyam, Sastry, Askell, et~al.]{brown2020language}
Tom Brown, Benjamin Mann, Nick Ryder, Melanie Subbiah, Jared~D Kaplan, Prafulla
  Dhariwal, Arvind Neelakantan, Pranav Shyam, Girish Sastry, Amanda Askell,
  et~al.
\newblock Language models are few-shot learners.
\newblock \emph{Advances in neural information processing systems},
  33:\penalty0 1877--1901, 2020.

\bibitem[Chiang et~al.(2023)Chiang, Li, Lin, Sheng, Wu, Zhang, Zheng, Zhuang,
  Zhuang, Gonzalez, Stoica, and Xing]{vicuna2023}
Wei-Lin Chiang, Zhuohan Li, Zi Lin, Ying Sheng, Zhanghao Wu, Hao Zhang, Lianmin
  Zheng, Siyuan Zhuang, Yonghao Zhuang, Joseph~E. Gonzalez, Ion Stoica, and
  Eric~P. Xing.
\newblock Vicuna: An open-source chatbot impressing gpt-4 with 90\%* chatgpt
  quality, 2023.

\bibitem[Chowdhery et~al.(2022)Chowdhery, Narang, Devlin, Bosma, Mishra,
  Roberts, Barham, Chung, Sutton, Gehrmann, et~al.]{chowdhery2022palm}
Aakanksha Chowdhery, Sharan Narang, Jacob Devlin, Maarten Bosma, Gaurav Mishra,
  Adam Roberts, Paul Barham, Hyung~Won Chung, Charles Sutton, Sebastian
  Gehrmann, et~al.
\newblock Palm: Scaling language modeling with pathways.
\newblock \emph{arXiv preprint arXiv:2204.02311}, 2022.

\bibitem[Chung et~al.(2022)Chung, Hou, Longpre, Zoph, Tay, Fedus, Li, Wang,
  Dehghani, Brahma, et~al.]{chung2022scaling}
Hyung~Won Chung, Le Hou, Shayne Longpre, Barret Zoph, Yi Tay, William Fedus,
  Eric Li, Xuezhi Wang, Mostafa Dehghani, Siddhartha Brahma, et~al.
\newblock Scaling instruction-finetuned language models.
\newblock \emph{arXiv preprint arXiv:2210.11416}, 2022.

\bibitem[Dai et~al.(2023)Dai, Li, Li, Tiong, Zhao, Wang, Li, Fung, and
  Hoi]{dai2023instructblip}
Wenliang Dai, Junnan Li, Dongxu Li, Anthony Meng~Huat Tiong, Junqi Zhao,
  Weisheng Wang, Boyang Li, Pascale Fung, and Steven Hoi.
\newblock Instructblip: Towards general-purpose vision-language models with
  instruction tuning, 2023.

\bibitem[Dong et~al.(2022)Dong, Li, Dai, Zheng, Wu, Chang, Sun, Xu, and
  Sui]{dong2022survey}
Qingxiu Dong, Lei Li, Damai Dai, Ce Zheng, Zhiyong Wu, Baobao Chang, Xu Sun,
  Jingjing Xu, and Zhifang Sui.
\newblock A survey for in-context learning.
\newblock \emph{arXiv preprint arXiv:2301.00234}, 2022.

\bibitem[Dong et~al.(2023)Dong, Han, Peng, Qi, Ge, Yang, Zhao, Sun, Zhou, Wei,
  et~al.]{dong2023dreamllm}
Runpei Dong, Chunrui Han, Yuang Peng, Zekun Qi, Zheng Ge, Jinrong Yang, Liang
  Zhao, Jianjian Sun, Hongyu Zhou, Haoran Wei, et~al.
\newblock Dreamllm: Synergistic multimodal comprehension and creation.
\newblock \emph{arXiv preprint arXiv:2309.11499}, 2023.

\bibitem[Gupta and Kembhavi(2023)]{gupta2023visual}
Tanmay Gupta and Aniruddha Kembhavi.
\newblock Visual programming: Compositional visual reasoning without training.
\newblock In \emph{Proceedings of the IEEE/CVF Conference on Computer Vision
  and Pattern Recognition}, pages 14953--14962, 2023.

\bibitem[Hao et~al.(2023)Hao, Liu, Wang, and Hu]{hao2023toolkengpt}
Shibo Hao, Tianyang Liu, Zhen Wang, and Zhiting Hu.
\newblock Toolkengpt: Augmenting frozen language models with massive tools via
  tool embeddings.
\newblock \emph{arXiv preprint arXiv:2305.11554}, 2023.

\bibitem[Ho et~al.(2022)Ho, Schmid, and Yun]{ho2022large}
Namgyu Ho, Laura Schmid, and Se-Young Yun.
\newblock Large language models are reasoning teachers.
\newblock \emph{arXiv preprint arXiv:2212.10071}, 2022.

\bibitem[Koh et~al.(2023)Koh, Fried, and Salakhutdinov]{koh2023generating}
Jing~Yu Koh, Daniel Fried, and Ruslan Salakhutdinov.
\newblock Generating images with multimodal language models.
\newblock \emph{arXiv preprint arXiv:2305.17216}, 2023.

\bibitem[Lai et~al.(2023)Lai, Tian, Chen, Li, Yuan, Liu, and Jia]{lai2023lisa}
Xin Lai, Zhuotao Tian, Yukang Chen, Yanwei Li, Yuhui Yuan, Shu Liu, and Jiaya
  Jia.
\newblock Lisa: Reasoning segmentation via large language model.
\newblock \emph{arXiv preprint arXiv:2308.00692}, 2023.

\bibitem[Latombe(2012)]{latombe2012robot}
Jean-Claude Latombe.
\newblock \emph{Robot motion planning}.
\newblock Springer Science \& Business Media, 2012.

\bibitem[Li et~al.(2022)Li, Li, Xiong, and Hoi]{li2022blip}
Junnan Li, Dongxu Li, Caiming Xiong, and Steven Hoi.
\newblock Blip: Bootstrapping language-image pre-training for unified
  vision-language understanding and generation.
\newblock In \emph{International Conference on Machine Learning}, pages
  12888--12900. PMLR, 2022.

\bibitem[Li et~al.(2023)Li, Li, Savarese, and Hoi]{li2023blip}
Junnan Li, Dongxu Li, Silvio Savarese, and Steven Hoi.
\newblock Blip-2: Bootstrapping language-image pre-training with frozen image
  encoders and large language models.
\newblock \emph{arXiv preprint arXiv:2301.12597}, 2023.

\bibitem[Liu et~al.(2023{\natexlab{a}})Liu, Li, Wu, and Lee]{liu2023llava}
Haotian Liu, Chunyuan Li, Qingyang Wu, and Yong~Jae Lee.
\newblock Visual instruction tuning.
\newblock In \emph{NeurIPS}, 2023{\natexlab{a}}.

\bibitem[Liu et~al.(2023{\natexlab{b}})Liu, Li, Wu, and Lee]{liu2023visual}
Haotian Liu, Chunyuan Li, Qingyang Wu, and Yong~Jae Lee.
\newblock Visual instruction tuning.
\newblock \emph{arXiv preprint arXiv:2304.08485}, 2023{\natexlab{b}}.

\bibitem[Liu et~al.(2023{\natexlab{c}})Liu, He, Wang, Wang, Wang, Chen, Zhang,
  Lai, Yang, Li, Yu, et~al.]{liu2023internchat}
Zhaoyang Liu, Yinan He, Wenhai Wang, Weiyun Wang, Yi Wang, Shoufa Chen,
  Qinglong Zhang, Zeqiang Lai, Yang Yang, Qingyun Li, Jiashuo Yu, et~al.
\newblock Interngpt: Solving vision-centric tasks by interacting with chatbots
  beyond language.
\newblock \emph{arXiv preprint arXiv:2305.05662}, 2023{\natexlab{c}}.

\bibitem[Ma et~al.(2023)Ma, Han, Wang, and Zhang]{ma2023cephgpt}
Lei Ma, Jincong Han, Zhaoxin Wang, and Dian Zhang.
\newblock Cephgpt-4: An interactive multimodal cephalometric measurement and
  diagnostic system with visual large language model.
\newblock \emph{arXiv preprint arXiv:2307.07518}, 2023.

\bibitem[Moon et~al.(2023)Moon, Madotto, Lin, Nagarajan, Smith, Jain, Yeh,
  Murugesan, Heidari, Liu, et~al.]{moon2023anymal}
Seungwhan Moon, Andrea Madotto, Zhaojiang Lin, Tushar Nagarajan, Matt Smith,
  Shashank Jain, Chun-Fu Yeh, Prakash Murugesan, Peyman Heidari, Yue Liu,
  et~al.
\newblock Anymal: An efficient and scalable any-modality augmented language
  model.
\newblock \emph{arXiv preprint arXiv:2309.16058}, 2023.

\bibitem[Mu et~al.(2023)Mu, Zhang, Hu, Wang, Ding, Jin, Wang, Dai, Qiao, and
  Luo]{mu2023embodiedgpt}
Yao Mu, Qinglong Zhang, Mengkang Hu, Wenhai Wang, Mingyu Ding, Jun Jin, Bin
  Wang, Jifeng Dai, Yu Qiao, and Ping Luo.
\newblock Embodiedgpt: Vision-language pre-training via embodied chain of
  thought.
\newblock \emph{arXiv preprint arXiv:2305.15021}, 2023.

\bibitem[OpenAI(2023)]{chatgpt}
OpenAI.
\newblock Chatgpt ({Mar} 14 version) [large language model].
\newblock \url{6}, 2023.

\bibitem[Ouyang et~al.(2022)Ouyang, Wu, Jiang, Almeida, Wainwright, Mishkin,
  Zhang, Agarwal, Slama, Ray, et~al.]{ouyang2022training}
Long Ouyang, Jeffrey Wu, Xu Jiang, Diogo Almeida, Carroll Wainwright, Pamela
  Mishkin, Chong Zhang, Sandhini Agarwal, Katarina Slama, Alex Ray, et~al.
\newblock Training language models to follow instructions with human feedback.
\newblock \emph{Advances in Neural Information Processing Systems},
  35:\penalty0 27730--27744, 2022.

\bibitem[Parisi et~al.(2022)Parisi, Zhao, and Fiedel]{parisi2022talm}
Aaron Parisi, Yao Zhao, and Noah Fiedel.
\newblock Talm: Tool augmented language models.
\newblock \emph{arXiv preprint arXiv:2205.12255}, 2022.

\bibitem[Patil et~al.(2023)Patil, Zhang, Wang, and Gonzalez]{patil2023gorilla}
Shishir~G Patil, Tianjun Zhang, Xin Wang, and Joseph~E Gonzalez.
\newblock Gorilla: Large language model connected with massive apis.
\newblock \emph{arXiv preprint arXiv:2305.15334}, 2023.

\bibitem[Peng et~al.(2023)Peng, Wang, Dong, Hao, Huang, Ma, and
  Wei]{peng2023kosmos}
Zhiliang Peng, Wenhui Wang, Li Dong, Yaru Hao, Shaohan Huang, Shuming Ma, and
  Furu Wei.
\newblock Kosmos-2: Grounding multimodal large language models to the world.
\newblock \emph{arXiv preprint arXiv:2306.14824}, 2023.

\bibitem[Qin et~al.(2023)Qin, Liang, Ye, Zhu, Yan, Lu, Lin, Cong, Tang, Qian,
  et~al.]{qin2023toolllm}
Yujia Qin, Shihao Liang, Yining Ye, Kunlun Zhu, Lan Yan, Yaxi Lu, Yankai Lin,
  Xin Cong, Xiangru Tang, Bill Qian, et~al.
\newblock Toolllm: Facilitating large language models to master 16000+
  real-world apis.
\newblock \emph{arXiv preprint arXiv:2307.16789}, 2023.

\bibitem[Rae et~al.(2021)Rae, Borgeaud, Cai, Millican, Hoffmann, Song,
  Aslanides, Henderson, Ring, Young, et~al.]{rae2021scaling}
Jack~W Rae, Sebastian Borgeaud, Trevor Cai, Katie Millican, Jordan Hoffmann,
  Francis Song, John Aslanides, Sarah Henderson, Roman Ring, Susannah Young,
  et~al.
\newblock Scaling language models: Methods, analysis \& insights from training
  gopher.
\newblock \emph{arXiv preprint arXiv:2112.11446}, 2021.

\bibitem[Schick et~al.(2023)Schick, Dwivedi-Yu, Dess{\`\i}, Raileanu, Lomeli,
  Zettlemoyer, Cancedda, and Scialom]{schick2023toolformer}
Timo Schick, Jane Dwivedi-Yu, Roberto Dess{\`\i}, Roberta Raileanu, Maria
  Lomeli, Luke Zettlemoyer, Nicola Cancedda, and Thomas Scialom.
\newblock Toolformer: Language models can teach themselves to use tools.
\newblock \emph{arXiv preprint arXiv:2302.04761}, 2023.

\bibitem[Shen et~al.(2023)Shen, Song, Tan, Li, Lu, and
  Zhuang]{shen2023hugginggpt}
Yongliang Shen, Kaitao Song, Xu Tan, Dongsheng Li, Weiming Lu, and Yueting
  Zhuang.
\newblock Hugginggpt: Solving ai tasks with chatgpt and its friends in
  huggingface.
\newblock \emph{arXiv preprint arXiv:2303.17580}, 2023.

\bibitem[Silver et~al.(2017)Silver, Schrittwieser, Simonyan, Antonoglou, Huang,
  Guez, Hubert, Baker, Lai, Bolton, et~al.]{silver2017mastering}
David Silver, Julian Schrittwieser, Karen Simonyan, Ioannis Antonoglou, Aja
  Huang, Arthur Guez, Thomas Hubert, Lucas Baker, Matthew Lai, Adrian Bolton,
  et~al.
\newblock Mastering the game of go without human knowledge.
\newblock \emph{nature}, 550\penalty0 (7676):\penalty0 354--359, 2017.

\bibitem[Tang et~al.(2023)Tang, Deng, Lin, Han, Liang, and
  Sun]{tang2023toolalpaca}
Qiaoyu Tang, Ziliang Deng, Hongyu Lin, Xianpei Han, Qiao Liang, and Le Sun.
\newblock Toolalpaca: Generalized tool learning for language models with 3000
  simulated cases.
\newblock \emph{arXiv preprint arXiv:2306.05301}, 2023.

\bibitem[Taori et~al.(2023)Taori, Gulrajani, Zhang, Dubois, Li, Guestrin,
  Liang, and Hashimoto]{alpaca}
Rohan Taori, Ishaan Gulrajani, Tianyi Zhang, Yann Dubois, Xuechen Li, Carlos
  Guestrin, Percy Liang, and Tatsunori~B. Hashimoto.
\newblock Stanford alpaca: An instruction-following llama model.
\newblock \url{https://github.com/tatsu-lab/stanford_alpaca}, 2023.

\bibitem[Touvron et~al.(2023{\natexlab{a}})Touvron, Lavril, Izacard, Martinet,
  Lachaux, Lacroix, Rozi{\`e}re, Goyal, Hambro, Azhar, Rodriguez, Joulin,
  Grave, and Lample]{touvron2023llama}
Hugo Touvron, Thibaut Lavril, Gautier Izacard, Xavier Martinet, Marie-Anne
  Lachaux, Timoth{\'e}e Lacroix, Baptiste Rozi{\`e}re, Naman Goyal, Eric
  Hambro, Faisal Azhar, Aurelien Rodriguez, Armand Joulin, Edouard Grave, and
  Guillaume Lample.
\newblock Llama: Open and efficient foundation language models.
\newblock \emph{arXiv preprint arXiv:2302.13971}, 2023{\natexlab{a}}.

\bibitem[Touvron et~al.(2023{\natexlab{b}})Touvron, Martin, Stone, Albert,
  Almahairi, Babaei, Bashlykov, Batra, Bhargava, Bhosale,
  et~al.]{touvron2023llama2}
Hugo Touvron, Louis Martin, Kevin Stone, Peter Albert, Amjad Almahairi, Yasmine
  Babaei, Nikolay Bashlykov, Soumya Batra, Prajjwal Bhargava, Shruti Bhosale,
  et~al.
\newblock Llama 2: Open foundation and fine-tuned chat models.
\newblock \emph{arXiv preprint arXiv:2307.09288}, 2023{\natexlab{b}}.

\bibitem[Vemprala et~al.(2023)Vemprala, Bonatti, Bucker, and
  Kapoor]{vemprala2023chatgpt}
Sai Vemprala, Rogerio Bonatti, Arthur Bucker, and Ashish Kapoor.
\newblock Chatgpt for robotics: Design principles and model abilities.
\newblock \emph{Microsoft Auton. Syst. Robot. Res}, 2:\penalty0 20, 2023.

\bibitem[Wang et~al.(2023)Wang, Chen, Chen, Wu, Zhu, Zeng, Luo, Lu, Zhou, Qiao,
  et~al.]{wang2023visionllm}
Wenhai Wang, Zhe Chen, Xiaokang Chen, Jiannan Wu, Xizhou Zhu, Gang Zeng, Ping
  Luo, Tong Lu, Jie Zhou, Yu Qiao, et~al.
\newblock Visionllm: Large language model is also an open-ended decoder for
  vision-centric tasks.
\newblock \emph{arXiv preprint arXiv:2305.11175}, 2023.

\bibitem[Wang et~al.(2022)Wang, Wei, Schuurmans, Le, Chi, Narang, Chowdhery,
  and Zhou]{wang2022self}
Xuezhi Wang, Jason Wei, Dale Schuurmans, Quoc Le, Ed Chi, Sharan Narang,
  Aakanksha Chowdhery, and Denny Zhou.
\newblock Self-consistency improves chain of thought reasoning in language
  models.
\newblock \emph{arXiv preprint arXiv:2203.11171}, 2022.

\bibitem[Wei et~al.(2022)Wei, Wang, Schuurmans, Bosma, Xia, Chi, Le, Zhou,
  et~al.]{wei2022chain}
Jason Wei, Xuezhi Wang, Dale Schuurmans, Maarten Bosma, Fei Xia, Ed Chi, Quoc~V
  Le, Denny Zhou, et~al.
\newblock Chain-of-thought prompting elicits reasoning in large language
  models.
\newblock \emph{Advances in Neural Information Processing Systems},
  35:\penalty0 24824--24837, 2022.

\bibitem[Weiss et~al.(1978)Weiss, Kulikowski, Amarel, and
  Safir]{weiss1978model}
Sholom~M Weiss, Casimir~A Kulikowski, Saul Amarel, and Aran Safir.
\newblock A model-based method for computer-aided medical decision-making.
\newblock \emph{Artificial intelligence}, 11\penalty0 (1-2):\penalty0 145--172,
  1978.

\bibitem[Wu et~al.(2023{\natexlab{a}})Wu, Yin, Qi, Wang, Tang, and
  Duan]{wu2023visual}
Chenfei Wu, Shengming Yin, Weizhen Qi, Xiaodong Wang, Zecheng Tang, and Nan
  Duan.
\newblock Visual chatgpt: Talking, drawing and editing with visual foundation
  models.
\newblock \emph{arXiv preprint arXiv:2303.04671}, 2023{\natexlab{a}}.

\bibitem[Wu et~al.(2023{\natexlab{b}})Wu, Fei, Qu, Ji, and Chua]{wu2023next}
Shengqiong Wu, Hao Fei, Leigang Qu, Wei Ji, and Tat-Seng Chua.
\newblock Next-gpt: Any-to-any multimodal llm.
\newblock \emph{arXiv preprint arXiv:2309.05519}, 2023{\natexlab{b}}.

\bibitem[Yang et~al.(2023)Yang, Song, Li, Zhao, Ge, Li, and
  Shan]{yang2023gpt4tools}
Rui Yang, Lin Song, Yanwei Li, Sijie Zhao, Yixiao Ge, Xiu Li, and Ying Shan.
\newblock Gpt4tools: Teaching large language model to use tools via
  self-instruction, 2023.

\bibitem[Yao et~al.(2022)Yao, Zhao, Yu, Du, Shafran, Narasimhan, and
  Cao]{yao2022react}
Shunyu Yao, Jeffrey Zhao, Dian Yu, Nan Du, Izhak Shafran, Karthik Narasimhan,
  and Yuan Cao.
\newblock React: Synergizing reasoning and acting in language models.
\newblock \emph{arXiv preprint arXiv:2210.03629}, 2022.

\bibitem[Yao et~al.(2023{\natexlab{a}})Yao, Yu, Zhao, Shafran, Griffiths, Cao,
  and Narasimhan]{yao2023tree}
Shunyu Yao, Dian Yu, Jeffrey Zhao, Izhak Shafran, Thomas~L Griffiths, Yuan Cao,
  and Karthik Narasimhan.
\newblock Tree of thoughts: Deliberate problem solving with large language
  models.
\newblock \emph{arXiv preprint arXiv:2305.10601}, 2023{\natexlab{a}}.

\bibitem[Yao et~al.(2023{\natexlab{b}})Yao, Li, and Zhao]{yao2023beyond}
Yao Yao, Zuchao Li, and Hai Zhao.
\newblock Beyond chain-of-thought, effective graph-of-thought reasoning in
  large language models.
\newblock \emph{arXiv preprint arXiv:2305.16582}, 2023{\natexlab{b}}.

\bibitem[You et~al.(2023)You, Zhang, Gan, Du, Zhang, Wang, Cao, Chang, and
  Yang]{you2023ferret}
Haoxuan You, Haotian Zhang, Zhe Gan, Xianzhi Du, Bowen Zhang, Zirui Wang,
  Liangliang Cao, Shih-Fu Chang, and Yinfei Yang.
\newblock Ferret: Refer and ground anything anywhere at any granularity, 2023.

\bibitem[Zeng et~al.(2022)Zeng, Liu, Du, Wang, Lai, Ding, Yang, Xu, Zheng, Xia,
  et~al.]{zeng2022glm}
Aohan Zeng, Xiao Liu, Zhengxiao Du, Zihan Wang, Hanyu Lai, Ming Ding, Zhuoyi
  Yang, Yifan Xu, Wendi Zheng, Xiao Xia, et~al.
\newblock Glm-130b: An open bilingual pre-trained model.
\newblock \emph{arXiv preprint arXiv:2210.02414}, 2022.

\bibitem[Zhang et~al.(2023{\natexlab{a}})Zhang, Han, Zhou, Hu, Yan, Lu, Li,
  Gao, and Qiao]{zhang2023llama}
Renrui Zhang, Jiaming Han, Aojun Zhou, Xiangfei Hu, Shilin Yan, Pan Lu,
  Hongsheng Li, Peng Gao, and Yu Qiao.
\newblock Llama-adapter: Efficient fine-tuning of language models with
  zero-init attention.
\newblock \emph{arXiv preprint arXiv:2303.16199}, 2023{\natexlab{a}}.

\bibitem[Zhang et~al.(2023{\natexlab{b}})Zhang, Sun, Chen, Xiao, Shao, Zhang,
  Chen, and Luo]{zhang2023gpt4roi}
Shilong Zhang, Peize Sun, Shoufa Chen, Min Xiao, Wenqi Shao, Wenwei Zhang, Kai
  Chen, and Ping Luo.
\newblock Gpt4roi: Instruction tuning large language model on
  region-of-interest.
\newblock \emph{arXiv preprint arXiv:2307.03601}, 2023{\natexlab{b}}.

\bibitem[Zheng et~al.(2023{\natexlab{a}})Zheng, He, and Wang]{zheng2023minigpt}
Kaizhi Zheng, Xuehai He, and Xin~Eric Wang.
\newblock Minigpt-5: Interleaved vision-and-language generation via generative
  vokens.
\newblock \emph{arXiv preprint arXiv:2310.02239}, 2023{\natexlab{a}}.

\bibitem[Zheng et~al.(2023{\natexlab{b}})Zheng, He, and
  Wang]{zheng2023minigpt5}
Kaizhi Zheng, Xuehai He, and Xin~Eric Wang.
\newblock Minigpt-5: Interleaved vision-and-language generation via generative
  vokens, 2023{\natexlab{b}}.

\bibitem[Zhou et~al.(2022)Zhou, Sch{\"a}rli, Hou, Wei, Scales, Wang,
  Schuurmans, Cui, Bousquet, Le, et~al.]{zhou2022least}
Denny Zhou, Nathanael Sch{\"a}rli, Le Hou, Jason Wei, Nathan Scales, Xuezhi
  Wang, Dale Schuurmans, Claire Cui, Olivier Bousquet, Quoc Le, et~al.
\newblock Least-to-most prompting enables complex reasoning in large language
  models.
\newblock \emph{arXiv preprint arXiv:2205.10625}, 2022.

\end{thebibliography}
}

\clearpage


\maketitlesupplementary
\section{ControlLLM-ChatGPT}
\label{sec:cllm_chatgpt}

\begin{table*}[!t]
\renewcommand{\arraystretch}{1.2}
 \begin{tabularx}{\textwidth}{l|X}
 \Xhline{1.5pt}
 \multicolumn{1}{c|}{\textbf{Field}}
 & \multicolumn{1}{c}{\textbf{Description}}  \\ 
 \hline
description  & a brief summary of subtask. It gives some guidance on how to approach the problem for ToG. \\
 \hline
domains & the domain scope that tools required by this task fall into. It helps ToG narrow down the search space and find the most relevant and suitable tools for the subtask. We showcase some domains in Table~\ref{tab:tool_set}.\\
 \hline
args & the inputs that the user provides for this subtask. It is usually in the form of key-value pairs, where the key is the type of the argument, and the value is the resource path or text you want to use. For example, [\{``type": ``image", ``value": ``image\_1.png"\}, \{``type": ``text", ``value": ``remove the dog in the picture"\}]. \\
 \hline
 return & the expected output of the subtask. For example, the return is \{``image": ``$\left< \texttt{GEN}\right>$-\texttt{0}''\}, which means the expected output is an image and ``$\left< \texttt{GEN}\right>$-\texttt{0}" is just a temporary placeholder. \\
 \bottomrule
\end{tabularx} 
\caption{
\small{\textbf{The output protocol of task decomposition.} We elaborate on each field in the output of task decomposition.}
}
\label{tab:decomposition_protocol}
\end{table*}

\renewcommand{\arraystretch}{1.2}
\begin{table*}[!thb]
    \centering
    \begin{tabular}{|m{17cm}|}
       \hline
        The following is a friendly conversation between a human and an AI. The AI is professional and parses user input to several tasks with lots of specific details from its context. If the AI does not know the answer to a question, it truthfully says it does not know.
The AI assistant can parse user input to several tasks with JSON format as follows: \textless Solution\textgreater [{``description": task\_description, ``task": [task\_domain\_1, task\_domain\_2], ``id": task\_id, ``dep": dependency\_task\_id, ``args": [{``type": ``text", ``image" or audio, ``value": text, image\_url or \textless GEN\textgreater -dep\_id}], ``returns":[{``type": ``segmentation", ``value": ``\textless GEN\textgreater -task\_id"}]}]\textless /Solution\textgreater .
The "description" should describe the task in detail, and AI assistant can add some details to improve the user’s request without changing the user’s original intention. The special tag ``\textless GEN\textgreater -dep\_id" refers to the one generated text/image/audio/video/segmentation/mask in the dependency task (Please consider whether the dependency task generates resources of this type.) and ``dep\_id" must be in ``dep" list. The special tag ``\textless GEN\textgreater -task\_id" refers to the one generated text/image/audio/video/segmentation/mask in this task and ``task\_id" should be in line with field ``id" of this task. The ``dep" field denotes the ids of the previous prerequisite tasks, which generate a new resource that the current task relies on. The ``args" field and the ``returns" field denotes the input resources and output resources of this task, respectively. The type of resource must be in [``text", ``image", ``line", ``normal", ``hed", ``scribble", ``pose", ``edge", ``bbox", ``category", ``segmentation", ``audio", ``video", ``segmentation", ``mask"], nothing else. The ``task" MUST be selected from the following options: ``question-answering", ``visual-question-answering", ``image-generation", ``image-editing", ``image-perception", ``image-processing", ``audio-perception", ``audio-generation", ``audio-editing", ``video-question-answering", ``video-perception", ``video-generation", ``video-editing", nothing else. Think step by step about all the tasks that can resolve the user's request.
Parse out as few tasks as possible while ensuring that the user request can be resolved. Pay attention to the dependencies and order among tasks. If some inputs of tools are not found, you cannot assume that they already exist. You can think of a new task to generate those args that do not exist or ask for the user's help. If the user request can't be parsed, you need to reply empty JSON [].
You should always respond in the following format:

\textless Solution\textgreater  \textless YOUR\_SOLUTION\textgreater  \textless /Solution\textgreater 

\textless YOUR\_SOLUTION\textgreater  should be strict with JSON format described above.\\
        \hline
    \end{tabular}
    \caption{\textbf{The prompt for task decomposition.} It is inspired by~\cite{shen2023hugginggpt}. }
    \label{tab:prompt_decomposition}
\end{table*}

In this variant, we implement language model $\mathcal{M}$ with ChatGPT. As such, we elaborately design a series of prompts for each module.

\subsection{Task Decomposition}
The prompt in Table~\ref{tab:prompt_decomposition} is designed for task decomposition in ControlLLM-ChatGPT. It guides the ChatGPT to decompose the user request into several subtasks.
Table~\ref{tab:decomposition_protocol} shows the output protocol of task decomposition.

\subsection{Tool Assessment}
\label{sec:tool_assessment}
Table~\ref{tab:prompt_tool} outlines a prompt design for the task of tool assessment, where the AI assistant evaluates tools' suitability for a given task. The output is captured in the JSON format, including reasons and a score. The scoring criteria range from 1 to 5, reflecting the tool's relevance to the task. The prompt emphasizes the connection between tool descriptions and task requirements. This prompt guides AI in making informed decisions when assessing tools' utility for a specific task.


\renewcommand{\arraystretch}{1.2}
\begin{table*}[!t]
    \centering
    \begin{tabular}{|m{17cm}|}
       \hline
        Given a task and a tool, the AI assistant helps the system decide whether this tool can process the task. The assistant should focus more on the description of the model and give a score to each tool.
The AI assistant respond with JSON format as follows: \textless Solution\textgreater {``Thought": {{thought}}, ``Score": {{score}} }\textless /Solution\textgreater .
The ``Thought" field records the model’s thinking process step by step within 80 words, which gives the reasons why giving this score.
The ``Score" field denotes a score that assesses whether this tool is useful for this task. Score is in [1, 2, 3, 4, 5]. Here are the scoring criteria: ``Score"=1: The tool is totally not related to the task and does not provide any useful output for solving the task. ``Score"=2: The tool is somewhat not related to the task and may not provide any useful output for solving the task. ``Score"=3: The tool is probably related to the task and provides some intermediate output that is partially helpful for solving the task, but it may not be the optimal one. ``Score"\textgreater 3: The tool is closely or directly related to the task and provides an output that is mostly helpful for solving the task or that matches the returns of the task with regard to the type. In a nutshell, for the given task, the higher the score, the more useful the tool is.
You should always respond in the following format: \textless Solution\textgreater  {{SOLUTION}} \textless /Solution\textgreater. $\backslash$n`SOLUTION` should strictly comply with the SON format described above. 
Task description: ``\{\{task\}\}".$\backslash$n$\backslash$n Here is the description of the tool ``\{\{tool\_name\}\}": $\backslash$n\{\{tool\_name\}\}: \{\{tool\_description\}\}$\backslash$nArgs: $\backslash$n\{\{arguments\}\}$\backslash$nReturns: $\backslash$n\{\{returns\}\}$\backslash$n$\backslash$nThe above information may be useful for AI to make decision. Please refer to the scoring criteria and score the tool \{\{tool\_name\}\} for this task. Notice that If the tool description contains keywords from the task description, the score of this tool should be greater than or equal to 3.\\
        \hline
    \end{tabular}
    \caption{\textbf{The prompt for tool assessment.}}
    \label{tab:prompt_tool}
\end{table*}
\subsection{Solution Expert}
\label{sec:solution_expert}
In this section, we delve into the core concept of the solution expert that streamlines the process of evaluating and selecting optimal solutions from all possible candidates. By systematically converting each solution into a formatted string description, the solution expert enables us to make informed decisions based on evaluated scores.

\textbf{Solution Description Formatting.}
To facilitate the solution expert to comprehend the solution, we need to generate the description for each solution candidate. This involves transforming raw solution data into structured, formatted string descriptions. These descriptions encapsulate the information including functionality, inputs and output.

\textbf{Solution Evaluation.}
The solution expert capitalizes on prompt engineering techniques to assess each solution based on subtask descriptions and formatted solution descriptions. The designed prompts guide language model $\mathcal{M}$ to evaluate the feasibility of each solution against the objective of the subtask. Through this process, we can assign scores to solutions, gauging their effectiveness and relevance to the task. It must ensure that the evaluation process is focused, targeted, and aligned with the subtask. The prompt template is shown in the Table~\ref{tab:prompt_solution}.

\textbf{Solution Ranking.}
The final aim of this module is to select the top-performing solutions. The optimal solution is identified as the highest score assessed in the solution evaluation. Given that sometimes the selected optimal solution may not meet the user requirements, we also provide several alternative solutions by setting a threshold score of 3. These solutions, which exhibit a higher degree of alignment with the subtask's requirements, emerge as the most promising candidates for user preference.

Through collaborative efforts, the optimal solution expert ensures that solutions are appropriately tailored, optimized, and well-adapted to the task.

\subsection{Resource Expert}
\label{sec:resource_expert}
In the algorithm of ToG, we encounter a challenge stemming from the potential presence of multiple instances of the same resource type within the available resource list. This challenge introduces complexity, making it difficult to straightforwardly deduce certain arguments for tools using predefined rules. As a result, we design a solution expert.

This module transforms the task of argument assignment into a fill-in-the-blank exercise.
To achieve this, we design a resource expert crafts with prompts that not only incorporate the task description but also include the available resource list. In this manner, a language model $\mathcal{M}$ is employed to dynamically complete the missing parameters within a solution by interacting with the contextual information presented. We put the prompt template in the Table~\ref{tab:prompt_resource}.

\renewcommand{\arraystretch}{1.2}
\begin{table*}[!t]
    \centering
    \begin{tabular}{|m{17cm}|}
       \hline
       Given a task and a solution, The AI assistant needs to score the solution and respond in JSON format. Please notice that the AI assistant should think. The AI assistant should pay more attention to the relevance between the description of each tool in the solution and task.
The AI assistant respond with JSON format as follows: \textless Solution\textgreater \{``Thought": ``thought", ``Score": score\}\textless /Solution\textgreater .
``Thought" field records the model’s thinking process step by step within 80 words, which gives the reasons why giving this score.
``Score" field denotes a score that assesses whether this tool is helpful for this task. ``Score" is in [1, 2, 3, 4, 5]. Here are the scoring criteria: ``Score"=1: The solution is totally not related to the user's request and can not solve the task. ``Score"=2: The solution is somewhat not related to the user's request and may not solve the task. ``Score"=3: The solution is probably related to the user's intention and may solve the task, but it may not be the optimal one. ``Score"\textgreater 3: The solution is closely or directly related to what the user wants and could satisfactorily solve the task. In a nutshell, the higher the score, the greater the likelihood of the solution solving the given task.
You should always respond in the following format: $\backslash$n\textless Solution\textgreater  `SOLUTION` \textless /Solution\textgreater $\backslash$n`SOLUTION` should strictly comply with the JSON format described above. $\backslash$nUser's request: ``\{\{request\}\}"
Task description: ``\{\{task\}\}".
Here is the description of the solution:
\{\{solution\}\}
Please refer to the scoring criteria and score this solution based on the task description. You should think carefully before scoring the solution. Notice that If the keywords in the solution are close in meaning to the keywords in the task description, then the score of this solution is at least 3. \\
        \hline
        
    \end{tabular}
    \caption{\textbf{The prompt for solution expert.}}
    \label{tab:prompt_solution}
\end{table*}

\renewcommand{\arraystretch}{1.2}
\begin{table*}[!t]
    \centering
    \begin{tabular}{|m{17cm}|}
       \hline
       Some tools have missing input arguments, and the AI assistant needs to infer these missing inputs from the context. Please notice that the AI assistant should never fake the resources that do not exist.
The returned input arguments should be JSON format as follows: [\{``image": ``xxx.png"\}, \{``bbox": ``\textless GEN\textgreater -detr-bbox-0"\}, ``text": ``\textless text\textgreater"]. AI assistant should always respond in the following format: $\backslash$n``\textless Explanation\textgreater  [briefly explain your choice here]\textless /Explanation\textgreater 
\textless Solution\textgreater  `SOLUTION` \textless /Solution\textgreater ". $\backslash$n`SOLUTION` should be strictly in the JSON format described above. $\backslash$nUser's request: $\backslash$n``\{\{request\}\}"$\backslash$nTask: $\backslash$n``\{\{task\_description\}\}". $\backslash$n\textless Resources\textgreater: $\backslash$n\{\{resources\}\}. $\backslash$nWe use \{\{tool\_name\}\} to solve this task: $\backslash$n`\{\{tool\_name\}\}`: \{\{tool\_description\}\} $\backslash$nArgs: 
\{\{arguments\}\} $\backslash$nReturns: 
\{\{returns\}\}
$\backslash$nFor the type of ``text", the AI assistant should summarize the content from the context based on the task and the tool's description. For other types of input, the AI assistant needs to select the inputs from \textless Resources\textgreater. Now we prepare the inputs for \{\{tool\_name\}\}: \{\{input\}\}. Please complete these inputs and return the completed inputs with the format described above like: \textless Solution\textgreater  `SOLUTION` \textless /Solution\textgreater . \\
       \hline
    \end{tabular}
    \caption{\textbf{The prompt for resource expert.}}
    \label{tab:prompt_resource}
\end{table*}

\renewcommand{\arraystretch}{1.2}
\begin{table*}[!t]
    \centering
    \begin{tabular}{|m{17cm}|}
       \hline
       Your name is ControlLLM, an AI-powered assistant. For user's request, the system executes the solution and collects the results based on the following workflow. You need to respond to user requests based on the following information. 
Here are the information for your reference.

\#\# User Request

\{\{request\}\}

\#\# Workflow and Execution Results

\{\{solution\}\}

\#\# Summarized Results

\{\{results\}\}

You must first answer the user’s request in a straightforward manner. Some of the results may not be accurate and need you to use your judgment in making decisions. Then please explain your workflow, including the tools and returned results for the request, in a friendly way. If the answers contain file paths, you have to repeat the complete file path. Only if there is nothing in the Summarized Results, you need to tell the user you can not finish the task.  \\
       \hline
    \end{tabular}
    \caption{\textbf{The Prompt Design in Response Generation.} We here refer to the prompts from~\cite{shen2023hugginggpt} to generate a user-friendly response.}
    \label{tab:prompt_response}
\end{table*}

\section{ControlLLM-LLaMA}
\label{sec:cllm_llama}
For ControlLLM-LLaMA, we use the LLaMA-7b~\cite{touvron2023llama} as language model $\mathcal{M}$ to solve the problems in task decomposition, tool assessment, solution expert, resource expert.
\label{appendix:lm}
\subsection{Instruction Generation} 

The first step to train $\mathcal{M}$ is to construct the instruction corpus. We here opt for ChatGPT-3.5 to generate the training corpus. 
The following steps will elaborate on the details of instructions generation for task decomposition, tool assessment, solution expert, and resource expert, respectively.

For task decomposition, we generate two different types of instructions as follows:
1) Basic instructions, where they only contain one subtask after task decomposition. We set some seed instructions with ground-truth results of task decomposition, which serve as initial templates for generating more diverse instructions. Then, we use ChatGPT to generate more diverse instructions based on the pre-defined seed instructions. During the generation process, we center on the seed instructions and produce more instructions using more diverse expressions and styles. These instructions need to share the task decomposition results with the seed instructions as ground truth.
2) Compound instructions, which involve multiple subtasks and intermediate resources. We simply assemble the basic instructions into the compound instructions in a coherent and logical manner. It aims to enhance the improve the complex interaction capability of the system by enabling the model to handle user requests that span multiple domains and require multiple steps of processing. We here generate almost 100k instructions for training. The instructions generated in this step will be used in the following tasks as well.

For the tool assessment, solution expert, and resource expert, 
we use prompts in Table~\ref{tab:prompt_tool}, ~\ref{tab:prompt_solution} and ~\ref{tab:prompt_resource} to collect the output from ChatGPT by running ControlLLM on the instructions generated above. Unlike directly generating the solution, these tasks only involve making a decision, like scoring the tools or solutions based on the input, so they are relatively simple, and ChatGPT with strong zero-shot capabilities, can easily solve them. Therefore, we opt to directly distill the knowledge of ChatGPT by using prompt techniques. Through the experiments, we verify the feasibility of this strategy.

\subsection{Training Recipes} 
We follow the training protocol in~\cite{alpaca}, where LLaMA~\cite{touvron2023llama} is used as an alternative choice for our language model $\mathcal{M}$. It is finetuned for three epochs with the initial learning rate 2e-5 and consine decay. We fix the training batch size as 128 by adaptively setting ``gradient\_accumulation\_steps". The whole training procedure is on 8xA100 GPUs.

\section{ControlLLM-Mix}
\label{sec:cllm_mix}
In practice, we find ControlLLM-ChatGPT has difficulty in task decomposition, especially for hard instructions. In addition, ControlLLM-LLaMA is good at task decomposition due to an instruction-tuned language model $\mathcal{M}$ while other modules are slightly inferior to ChatGPT. Because we finetune $\mathcal{M}$ by distilling the knowledge from ChatGPT to assess tools, ranking solutions, and assign arguments. As a result, we design ControlLLM-Mix to integrate the benefits from the other variants. For task decomposition, we use the same method from ControlLLM-LLaMA to finetune LLaMA-7B to decompose user requests. For the remaining modules, we directly utilize the ChatGPT, sharing the same prompt design from ControlLLM-ChatGPT. In experiments, ControlLLM-Mix achieves the best performance.

\section{Response Generation}
We design a prompt template for the Response Generation task in Table~\ref{tab:prompt_response}. In this task, the AI assistant is tasked with explaining the process and outcomes using input and inference results. The AI is instructed to respond directly to the user's request, followed by describing the task's procedure, offering analysis, and presenting model inference results using a first-person perspective. If the results involve file paths, the complete path should be provided, or if there are no results, the AI should communicate its inability. The prompt sets the context for generating informative and user-understandable responses during response generation.

\renewcommand{\arraystretch}{1.2}
\begin{table*}[!t]
 \begin{tabularx}{\textwidth}{l|X}
 \Xhline{1.5pt}
 \multicolumn{1}{c|}{\textbf{Domains}} & \multicolumn{1}{c}{\textbf{Tools}}\\ 
 \hline
question-answering & question\_answering, image\_question\_answering \\
\hline
natural-language-processing & summarization, title\_generation, text\_to\_tags, text\_to\_text\_generation, sentiment\_analysis \\
\hline
image-perception & object\_detection, image\_captioning, visual\_grounding, image\_classification, segment\_anything, instance\_segmentation, segment\_by\_points \\
\hline
image-generation & text\_to\_image, image\_to\_image, line\_text\_to\_image, hed\_text\_to\_image, scribble\_text\_to\_image, pose\_text\_to\_image, segmentation\_text\_to\_image, edge\_text\_to\_image, depth\_text\_to\_image, normal\_text\_to\_image\\
\hline
image-editing & text\_image\_editing, image\_inpainting, image\_cropping, mask\_image, highlight\_object\_on\_image \\
\hline
image-processing & image\_to\_edge, image\_to\_line, image\_to\_hed, image\_to\_scribble, image\_to\_pose, image\_to\_depth, image\_to\_normal \\
\hline
video-perception & video\_classification, video\_captioning \\
\hline
video-processing & dub\_video, video\_to\_webpage \\
\hline
video-generation & image\_audio\_to\_video, image\_to\_video, text\_to\_video\\
\hline
audio-perception & audio\_classification\\
\hline
audio-generation & text\_to\_music, text\_to\_speech, audio\_to\_Audio\\
 \bottomrule
\end{tabularx} 
\caption{
\small{\textbf{A subset of domains and tools in our tool kit.}}
}
\label{tab:tool_set}
\end{table*}

\section{Resource Types and Tools}
\label{sec:tool_type}
We initially define a type set containing a series of resource type descriptors, such as ``text", ``tags", ``html", ``image", ``video", ``audio", ``segmentation", ``edge", ``line", ``hed", ``canny", ``scribble", ``pose", ``depth", ``normal", ``mask", ``point", ``bbox" and ``category". The type set is easy to extend depending on the toolkit. The types of inputs of tools must fall into the pre-defined type set. 

In Table~\ref{tab:tool_set}, we exhibit lots of tools supported by our framework across different domains, including natural language, image, audio, video, \etc. The whole system will continue to evolve.

\renewcommand{\arraystretch}{1.2}
\begin{table*}[!tbh]
    \centering
    \begin{tabular}{|m{17cm}|}
       \Xhline{1.5pt}
        \multicolumn{1}{|C{17cm}|}{\textbf{Easy}}\\
        \hline
        1. Please determine if the image\_1.png contains a platyhelminth? \\
        2. How can I design a sleep monitoring system in C\# that can accurately detect a baby's specific sleep stages and predict when they will enter a light sleep stage within the next hour? And once this prediction is made, how can I alert the parent or caregiver that the baby will be waking up soon and suggest soothing methods to ease the transition from sleep to wakefulness? Also, how can I modify the statement ``The baby is sleeping" to reflect this predictive system in C\#? \\
        3. Please extract the scribble result for the image in image\_2.png" \\
        4. With the HED result image\_3.png, generate a new image that features a zoo landscape with various animals, a couple with their children, and a fountain.\\
        5. Given the image image\_4.png, What is unique about the window in the room? \\
        \hline
        \multicolumn{1}{|C{17cm}|}{\textbf{Medium}}\\
        \hline
        6. Can you generate a new image that has a similar layout to the file named 'image\_5.png'? I'm particularly interested in the positioning of the elements. The new image should have the same arrangement of elements and their positioning.\\
        7. Generate a new image conditioned on the segmentation from image\_6.png and the new image contains a majestic mountain range with a clear blue sky and a herd of wild horses running free. \\
        8. Take away the umbrella from the picture image\_7.png. \\
        9. Crop out the baseball glove in image\_8.png \\
        10. Provide me with a mask that separates the bear from the rest of the image\_9.png? \\
        \hline
        \multicolumn{1}{|C{17cm}|}{\textbf{Hard}}\\
        \hline
        11. provide the number of umbrellas presented in the image\_10.png, image\_11.png, image\_12.png, image\_13.png, image\_14.png \\
        12. Can you elaborate on the elements of the image\_15.png, image\_16.png and image\_17.png provided? \\
        13. Erase the laptop from the image\_18.png,image\_19.png and image\_20.png \\
        14. Create a new image using the segmentation from image\_21.png that showcases a cozy cabin in the woods with a dog and a cat, surrounded by snow-covered trees. Can you crop out the dog from given image? I'm looking for the cat in the image file, can you guide me to it? \\
        15. Can you determine whether image\_22.png contains a mouse? Please provide a list of all the objects present in the image, with a special emphasis on the killer. Is  image\_23.png displaying a banana? As for the image, what skills are important for improving one's performance in the depicted scenario?\\
        \Xhline{1.5pt}
    \end{tabular}
    \caption{\textbf{Test samples of instruction in the benchmark.}}
    \label{tab:instructions}
\end{table*}
\section{Metrics for Tool Selection} 
\label{sec:metric}

A) Irrelevant Tool Inclusion Rate (\textit{abbr.} $IR$): 
\begin{equation}
\begin{aligned}
F(s^p) = \left\{
\begin{aligned}
\text{true} &, \quad \text{$s^p$ contains the irrelevant tools} \\
\text{false} &,\quad \text{otherwise}
\end{aligned}
\right.,
\end{aligned}
\end{equation}
\begin{equation}
\begin{aligned}
IR = \frac{\sum_{i}^{|S^p|} \mathbb{I}(F(S_{i}^{p}))}{|S^p|},
\end{aligned}
\end{equation}
where $\mathbb{I}$ is indicator function, $\mathbb{|\cdot|}$ represents the number of elements in a set, and $S^p$ denotes all predicted solutions on our benchmark. It measures the proportion of the predicted solutions that contain the irrelevant tools. A higher $IR$ indicates that the method tends to include more unnecessary tools, potentially hindering effective task planning. This metric gauges the performance of methods in excluding irrelevant tools. 

B) Necessary Tool Inclusion Rate (\textit{abbr.} $NR$): 
\begin{equation}
\begin{aligned}
H(s^p) = \left\{
\begin{aligned}
\text{true} &,\quad \text{Solution $s^p$ contains necessary tools} \\
\text{false} &,\quad \text{otherwise}
\end{aligned}
\right.,
\end{aligned}
\end{equation}
\begin{equation}
\begin{aligned}
NR = \frac{\sum_{i}^{|S^p|} \mathbb{I}(H(S_{i}^{p}))}{|S^p|}.
\end{aligned}
\end{equation}
The necessary tools play a critical role in solving the user request. For example, if users want to know the position of a specific object, the object detection tool is necessary. This metric measures the proportion of solutions that contain the necessary tools for solving the task but without considering whether the arguments of tools are correct. It checks whether the solution has all the necessary tools that can produce the expected output. A high NR value means that the method has a strong ability in task planning and finding the right tools for the user’s request.

\subsection{Metrics for Argument Assignment}
A) Resource Hallucination Rate (\textit{abbr.} $HR$): 
\begin{equation}
\begin{aligned}
P(s^p) = \left\{
\begin{aligned}
\text{true} &,\quad s^p \text{contains false resources} \\
\text{false} &,\quad \text{otherwise}
\end{aligned}
\right.,
\end{aligned}
\end{equation}
\begin{equation}
\begin{aligned}
HR =& \frac{\sum_{i}^{|S^p|} \mathbb{I}(P(S_{i}^{p}))}{|S^p|}.
\end{aligned}
\end{equation}
This indicator reveals the extent of hallucination when inferring the arguments for tools. It checks whether all the arguments of the tools exist in the input resources or not. A low HR value means that the method rarely leads to hallucinations that are common in LLMs.

B) Resource Type Consistency Rate (\textit{abbr.} $CR$): 
\begin{equation}
\begin{aligned}
Q(s^p) = \left\{
\begin{aligned}
\text{true} &,\quad \text{No resource type conflict in $s^p$} \\
\text{false} &,\quad \text{otherwise}
\end{aligned}
\right.,
\end{aligned}
\end{equation}
\begin{equation}
\begin{aligned}
CR = \frac{\sum_{i}^{|S^p|} \mathbb{I}(Q(S_{i}^{p}))}{|S^p|}.
\end{aligned}
\end{equation}
This metric examines whether the types of resources used as inputs for the predicted solution match those of the corresponding tools. It evaluates the method's ability to ensure consistency between argument types and tools. A high CR value means that the method can correctly infer and assign arguments for each tool.

\subsection{Solution Evaluation}

The Solution Evaluation (\textit{abbr.} $SE$):
\begin{equation}
\begin{aligned}
W(s^p) = \left\{
\begin{aligned}
\text{true} &, \quad \text{$s^p$ can solve the task}\\
\text{false} &,\quad \text{otherwise}
\end{aligned}
\right.,
\end{aligned}
\end{equation}
\begin{equation}
\begin{aligned}
SE = \frac{\sum_{i}^{|S^p|} \mathbb{I}(W(S_{i}^{p}))}{|S^p|}.
\end{aligned}
\end{equation}
This metric measures the success rate of all generated solutions on our benchmark, regardless of whether it contains irrelevant tools, as long as the chain of tool invoking outputs the information that is able to solve the task. A higher score in the solution evaluation indicates that the method is able to provide an effective solution to user requests.

In summary, these intuitive metrics together provide a comprehensive assessment of tool-augmented LLMs in terms of tool selection, argument assignment, and overall effectiveness in addressing user queries.

\section{Case Studies}
\label{sec:case_studies}
\begin{figure*}[t]
\centering
\includegraphics[width=0.85\linewidth]{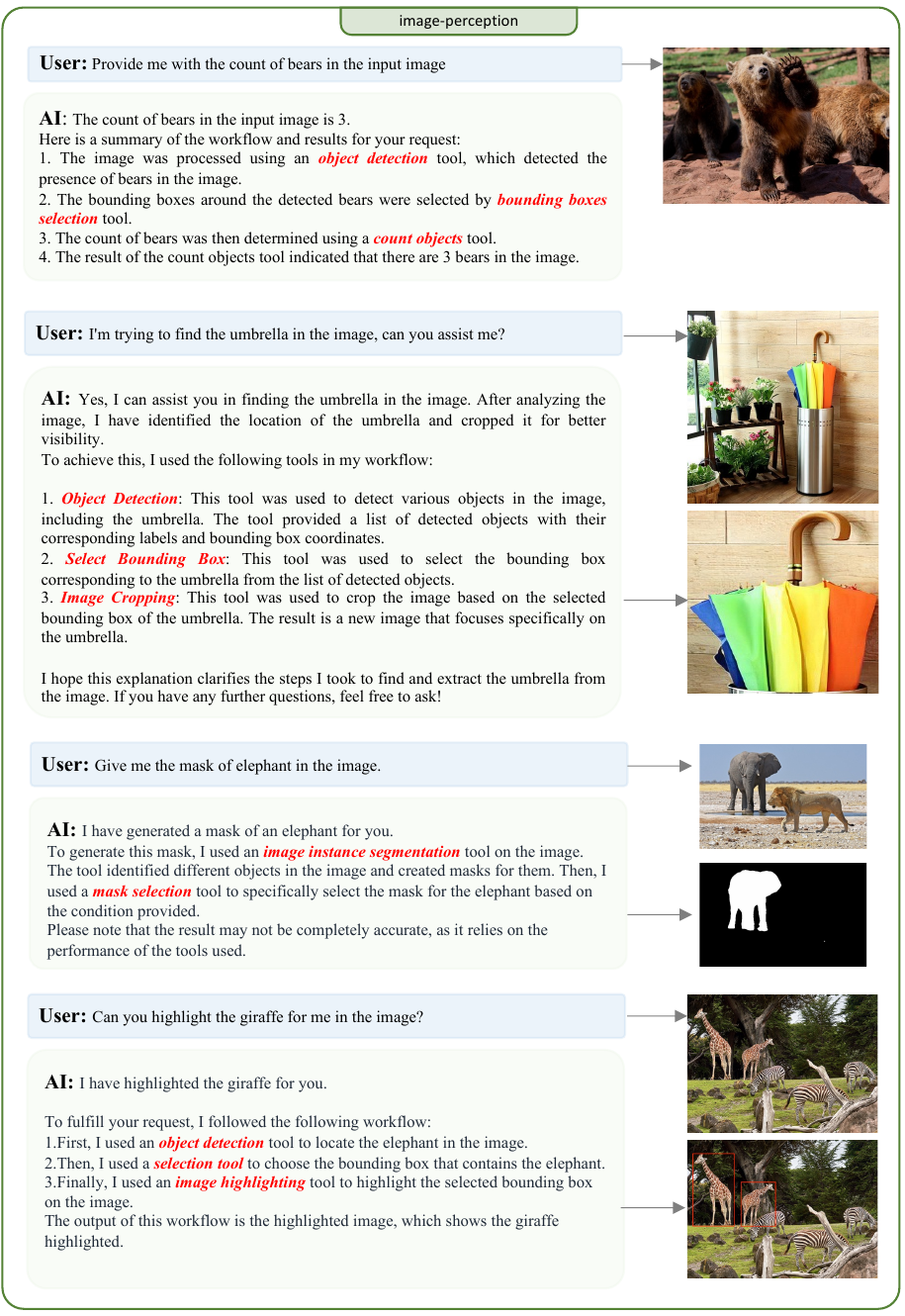}
\caption{\textbf{Case study on different task domains} (a).}
\label{fig:vis_1}
\end{figure*}

\begin{figure*}[t]
\centering
\includegraphics[width=0.85\linewidth]{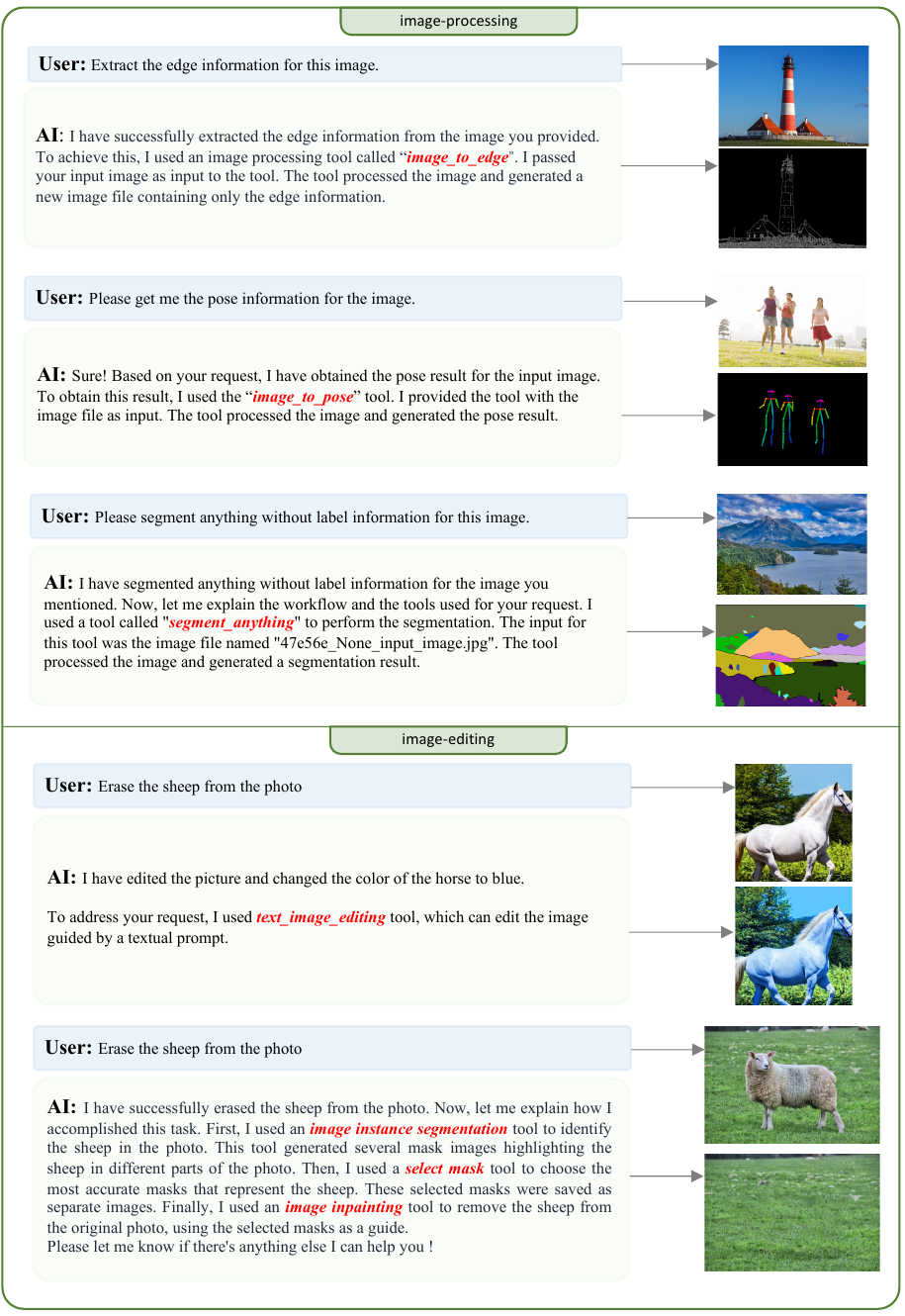}
\caption{\textbf{Case study on different task domains} (b).}
\label{fig:vis_2}
\end{figure*}

\begin{figure*}[t]
\centering
\includegraphics[width=0.85\linewidth]{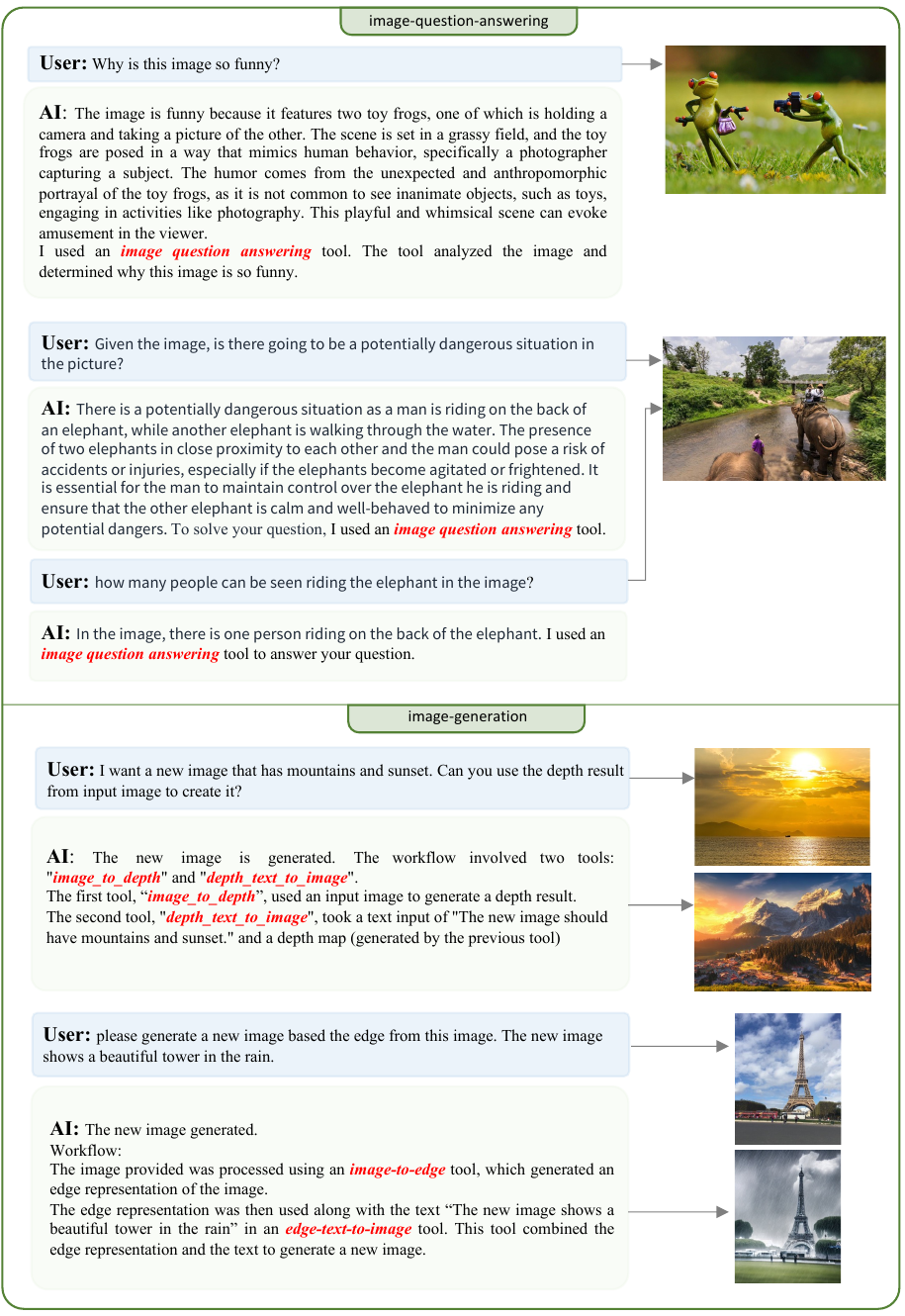}
\caption{\textbf{Case study on different task domains} (c).}
\label{fig:vis_3}
\end{figure*}

\begin{figure*}[t]
\centering
\includegraphics[width=0.85\linewidth]{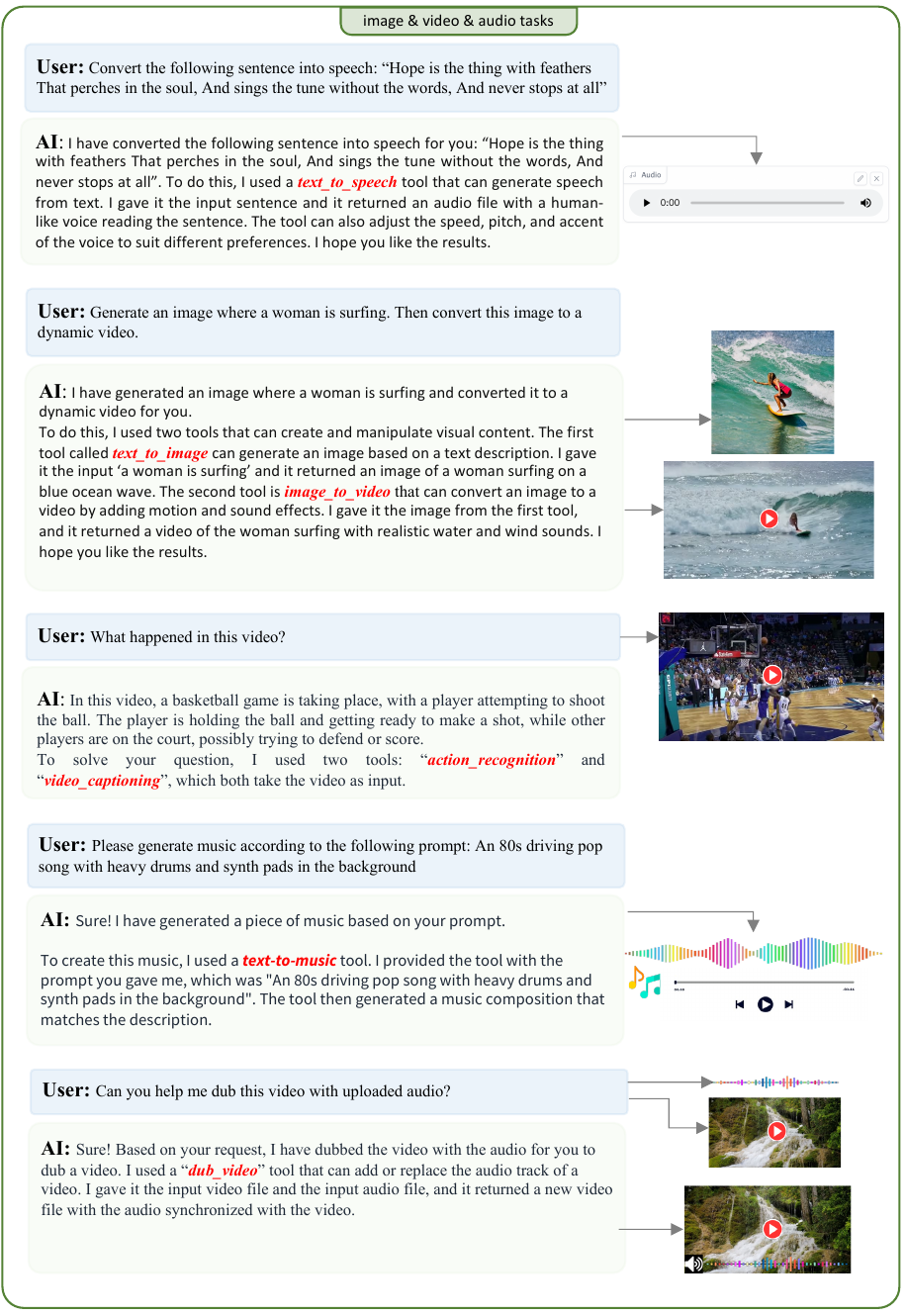}
\caption{\textbf{Case study on different task domains} (d).}
\label{fig:vis_4}
\end{figure*}

\begin{figure*}[t]
\centering
\includegraphics[width=0.85\linewidth]{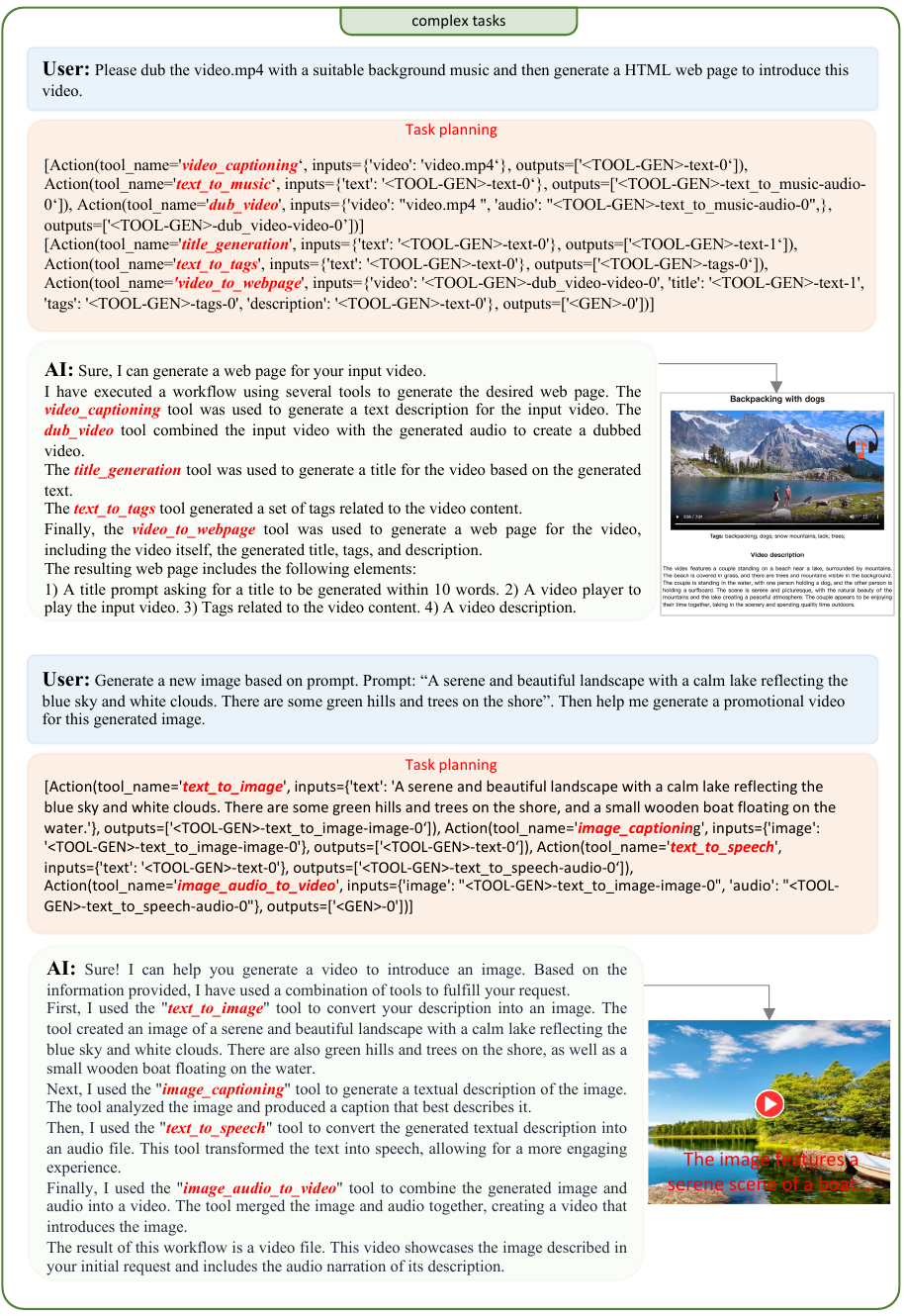}
\caption{\textbf{Case study on different task domains} (e).}
\label{fig:vis_5}
\end{figure*}

\begin{figure*}[t]
\centering
\includegraphics[width=0.85\linewidth]{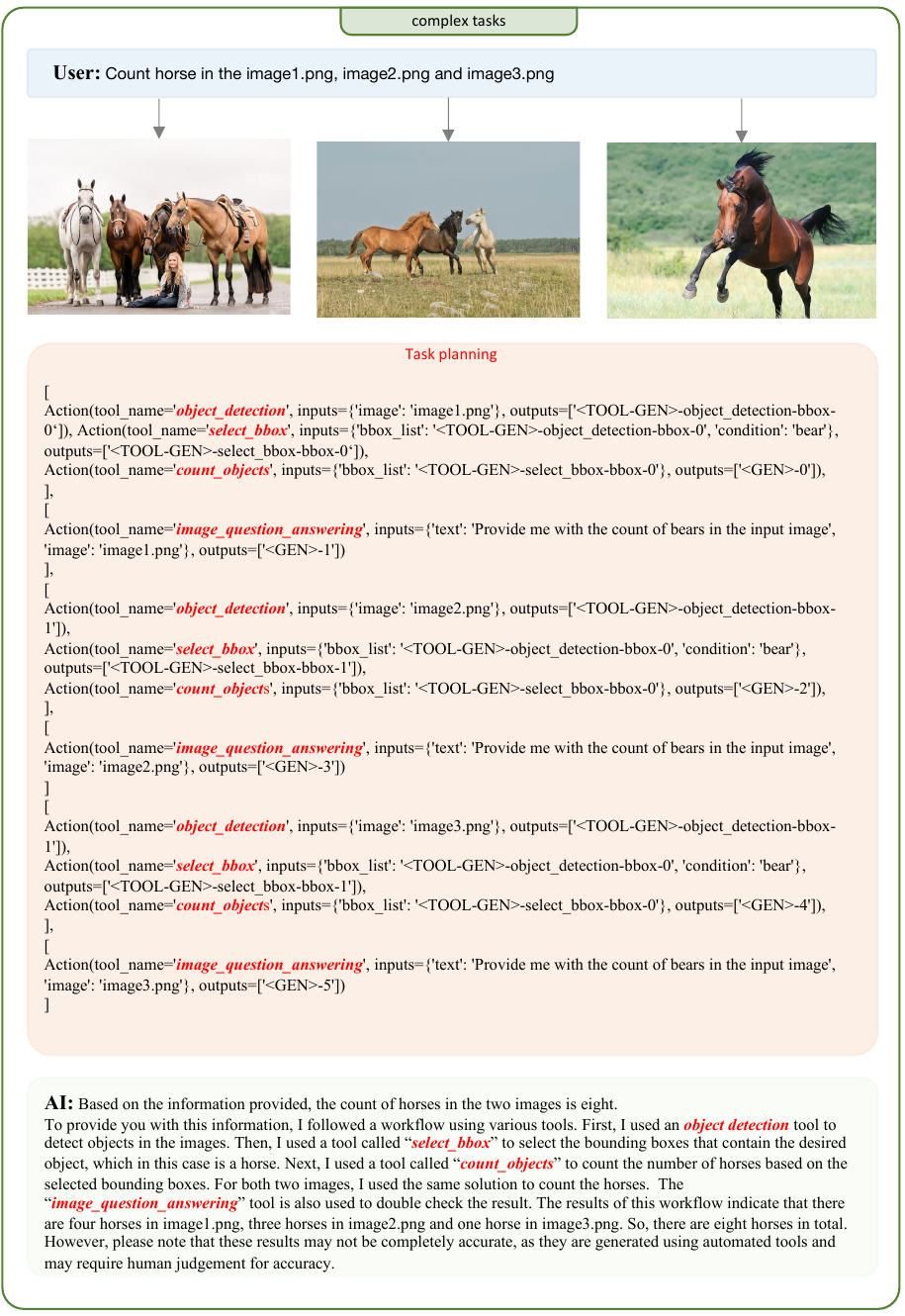}
\caption{\textbf{Case study on different task domains} (f).}
\label{fig:vis_6}
\end{figure*}

We provide more cases across different modalities to validate the user experience for our method in practice. In Fig.~\ref{fig:vis_1}, we show some cases of image perception, which involves analyzing and understanding the content of an image, such as detecting objects, counting objects, finding objects, segmenting objects, answering questions about the image, \etc. These tasks require the system to invoke tools to process visual information and extract relevant features and labels from the image.
Fig.~\ref{fig:vis_2} gives examples of image processing and image editing, which assist users in processing or editing the image according to some criteria or instructions.
Fig.~\ref{fig:vis_3} mainly focuses on image question answering and image generation, showing the graphic dialogue capability. In Fig.~\ref{fig:vis_4}, we provide some multi-modal interaction cases on image, video, and audio domains. In addition, we also illustrate the capabilities of complicated scenarios with solutions searched by ToG during task planning in Fig.~\ref{fig:vis_5} and Fig.~\ref{fig:vis_6}. These complex tasks involve combining multiple tools to find an advanced and creative solution path that can solve more challenging problems. It requires a system that can integrate different types of information and outputs from tools and generate comprehensive and meaningful responses based on execution results. These figures demonstrate the strong capabilities of ControlLLM in task planning for both simple and complicated scenarios. It thus leads to a better user experience.

\end{document}